\documentclass[fleqn,10pt]{wlscirep}
\usepackage[utf8]{inputenc}
\usepackage[T1]{fontenc}
\usepackage{multirow}
\usepackage{verbatim}
\usepackage{chngcntr}
\usepackage{makecell}
\usepackage{pdfpages}

\title{PointIso: Point Cloud Based Deep Learning Model for Detecting Arbitrary-Precision Peptide Features in LC-MS Map through Attention
Based Segmentation}

\author[1]{Fatema Tuz Zohora}
\author[2]{M Ziaur Rahman}

\author[1]{Ngoc Hieu Tran}
\author[2]{Lei Xin}
\author[2]{Baozhen Shan} 
\author[1,*]{Ming Li}

\affil[1]{David R. Cheriton School of Computer Science, University of Waterloo, Waterloo, ON N2L 3G1, Canada}
\affil[2]{Bioinformatics Solutions Inc., Waterloo, ON N2L 6J2, Canada}

\affil[*]{mli@uwaterloo.ca}


\begin{abstract}
A promising technique of discovering disease biomarkers is to measure the relative protein abundance in multiple biofluid samples through liquid chromatography with tandem mass spectrometry (LC-MS/MS) based quantitative proteomics. The key step involves peptide feature detection in LC-MS map, along with its charge and intensity. Existing heuristic algorithms suffer from inaccurate parameters since different settings of the parameters result in significantly different outcomes. Therefore, we propose PointIso, to serve the necessity of an automated system for peptide feature detection that is able to find out the proper parameters itself, and is easily adaptable to different types of datasets. It consists of an attention based scanning step for segmenting the multi-isotopic pattern of peptide features along with charge and a sequence classification step for grouping those isotopes into potential peptide features. PointIso is the first point cloud based, arbitrary-precision deep learning network to address the problem and achieves 98\% detection of high quality MS/MS identifications in a benchmark dataset, which is higher than several other widely used algorithms. Besides contributing to the proteomics study, we believe our novel segmentation technique should serve the general image processing domain as well.  
\end{abstract}

\begin{document}
%\pdfoutput=1
\flushbottom
\maketitle
%PointIso has two times higher input resolution and three times faster running speed than the first version.


\thispagestyle{empty}
\section*{Introduction}
Deep learning, an emerging branch from machine learning, has exhibited unprecedented performance in a wide range of research areas, including image processing, pattern recognition, natural language processing and many others~\cite{lecun2015deep}. Deep Learning is accelerating new scientific discoveries not only in core machine learning problems, but also many multidisciplinary sectors, like bioinformatics, autonomous driving, fraud detection,  etc. Proteomics is the large scale study of proteins - main workhorses responsible for biological functions and activities in a cell, tissue, or organism. Liquid chromatography coupled with tandem mass spectrometry (LC-MS/MS) based analysis leads the pathway of disease biomarker identification~\cite{jaffe2006pepper}, antibody sequencing~\cite{tran2016complete},  neoantigen detection\cite{bulik2019deep}, drug discovery and many other clinical research~\cite{aoshima2014simple}. Since peptide is the building block of protein, there has been significant effort on peptide identification, sequencing, and quantitation. Most of these problems involve pattern recognition and sequence prediction. That is why the popular deep learning models, e.g., convolutional neural network (CNN), recurrent neural network (RNN), etc. are worth applying in proteomics domain as well. Bulik-Sullivan et al.~\cite{bulik2019deep} proposed a deep learning based model EDGE, that shows improved ability to develop neoantigen-targeted immunotherapies for cancer patients using tumor HLA peptide mass spectrometry datasets.  Tran et al.~\cite{tran2019deep} proposed DeepNovo-DIA, a deep learning based de novo peptide sequencing technique. PROSIT is proposed by Gessulat et al.~\cite{gessulat2019prosit} that offers a proteome-wide prediction of peptide tandem mass spectra by deep learning.  Guan et al.~\cite{guan2019prediction} proposed deep learning strategy for the prediction of LC-MS/MS properties of peptides from sequence. AlphaFold~\cite{senior2020improved}, developed by Google’s DeepMind, makes significant progress on protein folding, one of the core challenges in biology. We proposed DeepIso~\cite{zohora2019deepiso}, that combines recent advances in CNN and RNN to detect peptide features of different charge states, as well as, estimate their intensity. It is the first deep learning based model to address the target problem and should be applicable in protein quantitation and biomarker discovery as well. However, it is a fixed precision model (up to 2 decimal places) and comparatively slower than other competitive tools. But it gave us a good insight on the scope of deep learning in this context. So we make  significant improvements in DeepIso and propose PointIso that resolves all those  shortages.

Many diseases are fundamentally linked to proteins. Therefore, if we can measure the relative protein abundance between the biofluid samples from healthy person and disease afflicted person, we can identify the proteins which are either diagnostic or prognostic of the disease. Such proteins are called disease biomarkers. The LC-MS/MS based analysis is the current state-of-the-art technology for protein identification and quantification~\cite{aebersold2003mass}. The procedure starts with digesting the protein into smaller peptides by various sequence-specific enzyme, and then the protein sample is passed to the first mass spectrometer (MS), where the peptides are ionized. Output of the first MS is called LC-MS map or MS1 data, that contains the three dimensional peptide features as shown in Figure~\ref{alg}. The three dimensions are: mass-to-charge ($m/z$ or Th), retention time (RT), and intensity (I) of peptide ions in that sample. %Usually the LC-MS map span over 400 to 2000 $m/z$ range of $m/z$  with very high resolution (as high as up to 4 decimal places) and up to 120 minutes experiment time.
Each peptide feature consists of multiple isotopes and appears during its elution time (RT range) in the map. %This pattern is formed by different molecular isotopes, e.g. carbon-12 and carbon-13, of the same peptide. Usually the first isotope has peak intensity and is called precursor ion. 
The precursor ions (peak intensity isotope in the feature) are further passed to the second MS which generates MS/MS fragmentation spectrum that facilitate the identification of peptide sequence~\cite{steen2004abc}, i.e., the amino acid sequence. %Peptide feature detection from LC-MS map is directly responsible for protein quantification.
%Label Free Quantification (LFQ) has the potential for identification and quantification of differentially expressed proteins in normal and diseased samples~\cite{atrih2014quantitative}, and are mostly applicable to the data acquired on mass spectrometers equipped with the time-of-flight (Tof), Fourier transform-ion cyclotron resonance (FT-LTQ), or OrbiTrap mass analyzers. This enables the extraction of peptide signals for specific analytes on the MS1 level and thus uncouples the quantification from the identification process. 
We have to find the precise peptide feature boundary from LC-MS map, since this is a crucial step for many downstream workflows, e.g., protein quantification, identification of chimeric spectra, biomarker identification, etc. %Therefore, we believe its worth investigating this problem using deep learning based pattern recognition techniques.

%Detecting multi-isotope patterns in LC-MS map is a  challenging task due to the overlapping peptides, several charges of the same molecule, and intensity variation. Moreover, a single LC-MS map may have gigapixel size containing thousands to millions of peptide features. 

There has been past attempts of peptide feature detection using several heuristic algorithms whose parameters are set by domain experts through rigorous experiments, and different dataset needs different set of parameters. Its prone to human error since different settings result in significantly different outcomes. Therefore, our target problem is to build up an automated system that learns the parameters itself using the power of deep learning  through several layers of neurons by training on appropriate dataset. To be specific, we propose a deep learning based model that  detects the peptide feature boundary along with it's charge state. Although this resemblance common pattern recognition problems in machine learning, however, 
there are several reasons what make the peptide feature detection far more challenging than the general cases. For instance, frequent overlapping among the multi-isotope patterns, %(illustrated later in Discussion section)
 not always obeying typical conventions, need for filtering out from feature like noisy signals, and hundreds of thousands of  peptide features who are comparatively tiny in size as compared to the massive background (RT axis span over 0 to 120 minutes, and $m/z$ axis ranges from 400 to 2000 $m/z$ with resolution as high as up to 4 decimal places). DeepIso~\cite{zohora2019deepiso} shows better performance than other existing heuristics based tools. However, it cannot accept very high resolution input due to using image based CNN, and %although its important to disclose important hidden features. 
comparatively slower than other competitive tools because of using classification network in a overlapping sliding window approach for doing the feature  segmentation. Therefore, we bring  significant changes to overcome these problems and offer PointIso that gives precise boundary information in a time efficient manner and achieves higher percentage of feature detection. In particular, we combine point cloud based deep neural network PointNet~\cite{qi2017pointnet} and Dual Attention Network (DANet)~\cite{fu2019dual} to integrate local features with their global dependencies and some context information. %image representation of 3D objects and takes longer time.
Point cloud is a data structure for representing objects using points, e.g., using triplets in a three dimensional environment. Unlike DeepIso where 2D projected image are used for representing 3D peptide features, we adapt PointNet to our context in order  to directly process the 3D features. It makes it feasible to accept input data with two or more times higher resolution (arbitrary-precision) than DeepIso and achieves better detection. On the other hand, the original DANet is proposed for finding the correlated objects in the input landscape image for the autonomous driving problem. %The first version of DeepIso converts the 3D peptide features to a 2D projected image for easier application of CNN and RNN. However, it makes the data  unnecessarily voluminous and causes time issues. 
 %Another notable contribution of this work is to combine DANet with PointNet during segmenting peptide features in the LC-MS map or MS1 data. 
 We take the idea and plug it into the PointNet network to solve boundary value problems during scanning the huge LC-MS map through \emph{non-overlapping} sliding windows, which makes it three times faster than DeepIso. Therefore, PointIso achieves a higher peptide feature detection rate with a faster speed than before.
% Please note that, overlapping sliding windows is not a wise option here since it would take weeks to finish due to the high resolution input. That is why the new concept deemed necessary to achieve a faster speed with high resolution input data.

Our novel concept of attention based scanning of LC-MS map through completely non-overlapping sliding window for peptide feature detection not only leads us to a high feature detection rate in a reasonable running time, but also has the potential to serve the general image processing problems. Besides that, our deep learning model is  easily adaptable to new cases by fine tuning through retraining with the misclassified samples. Therefore, we believe our research work discloses new research directions, as well as, makes a notable contribution in accelerating the progress of deep learning in proteomics study.

%Please note that the full LC-MS plot cannot be passed as a single input image due to its gigapixel size, 

%In most of the existing algorithms for peptide feature detection, many parameters are set based on experience with empirical experiments, whose different settings may have a large impact on the outcomes. In contrast to these existing works, our research aims at systematically training a deep neural network using real dataset to automatically learn all characteristics of the data, without human intervention. Last but not least, even if the model makes wrong predictions, the correct results can be put back as new training data so that the model can learn from its own mistakes. We believe that such models shall have superior performance over existing techniques and shall become the method of choice soon.

\section*{Results}
\begin{figure}[h!]
\centering
\includegraphics[scale = .57]{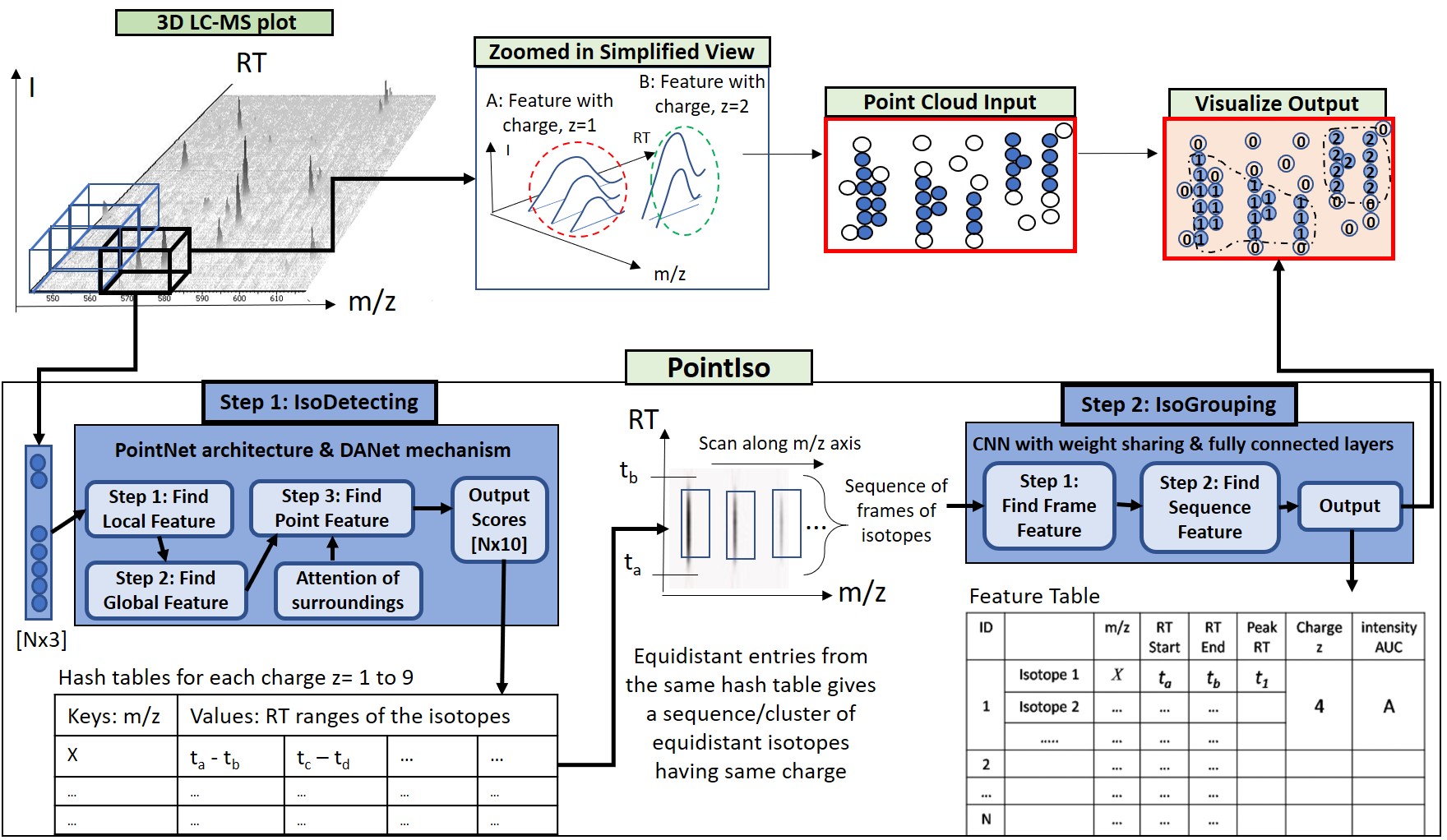}
\caption{Workflow of our proposed model PointIso to detect peptide features from LC-MS map of protein sample}\label{alg}
\end{figure}

%We first discuss about the datasets used in our experiments. Then the details on the evaluation strategy, along with the test results will be followed.  
We explain the intuition of our proposed model using the workflow shown in Figure~\ref{alg}. We see the three dimensional LC-MS map in the upper left corner and PointIso starts with scanning this map by sliding a window along two directions: $m/z$ axis and RT axis. A sliding window is essentially a 3D cube, which can be represented using point cloud. We show a random scanning window in bold black boundary, enclosing two features. This region is further shown in the next image, labeled as `Zoomed in Simplified View'. Here, two features A and A are shown using red and green boundary. The corresponding point cloud version of this window is shown in the next image, labeled as  `Point Cloud Input'. Here the blue and white points correspond to the features and background points respectively. We also see that PointIso model works through two modules, IsoDetecting in the first step, and IsoGrouping in the second step.
So the point cloud input consists of a set of `N' datapoints which is passed as input to the IsoDetecting module as shown by the arrow sign from the sliding 3D window. IsoDetecting module segments the datapoints as z = 0 to 9, where z=0 means the respective datapoint is a noise, and z = 1 to 9 means the respective datapoint belongs to a feature having charge z. We build this module by incorporating the attention mechanism offered by DANet into the PointNet architecture, in order to support \emph{non-overlapping} sliding windows. IsoDetecting module gives us a list of isotopes of potential features, and are recorded in a hash table.  %where the $m/z$ index of isotopes are used as key and RT range of those isotopes are inserted as value under those keys. 
Then in the second step, IsoGrouping module takes those sequences of isotopes (each sequence may have any number of isotopes) and predicts the boundary (first and last isotope) of features. Each sequence may be broken into multiple features or merely predicted as noisy signals. This prediction finally gives us a feature table that reports the detected peptide features along with the $m/z$ of monoisotope (the first isotope of a feature), charge, RT range of each isotope, and intensity. We can also visualize the final result as shown in the image labeled as `Visualize Output' in Figure~\ref{alg} (upper right corner).

We downloaded the benchmark dataset from ProteomeXchange (PXD001091) which was prepared by Chawade et al.~\cite{chawade2014data} for data-dependent acquisition (DDA). The samples consist of a long-range dilution series of synthetic peptides (115 peptides from potato and 158 peptides from human) spiked in a %total cell protein digest. 
background of stable and nonvariable peptides, obtained from \emph{Streptococcus pyogenes} strain SF370~\cite{teleman2012automated}. Synthetic peptides were spiked into the background at 12 different concentration points resulting 12 samples each having multiple replicates. We obtain LC-MS map from each replicate, totaling 57 LC-MS maps for the experiment.
\subsection*{Training of PointIso}
Since we are using supervised learning approach, we need labeled data for training. Human annotation of peptide features is out of scope due to gigapixel size of the LC-MS maps~\cite{tautenhahn2008highly}. Therefore, we match the feature lists produced by MaxQuant 1.6.3.3 and Dinosaur 1.1.3 with a tolerance of 10 ppm $m/z$ and 0.03 RT and take the common set as the training samples. %, as done in related  literature~\cite{tautenhahn2008highly, sturm2008openms, rost2016openms}. 
In PointIso we also need the precise boundary information (i.e., RT time range and $m/z$ value of each isotopes of the features) which is not generated for the users in MaxQuant. Therefore we use Dinosaurs for that information. The IsoDetecting and IsoGrouping modules are trained separately using suitable training data. In order to generate training samples for IsoDetecting module, we place scanning window over the features and cut the region along with surrounding area. The total number of features available from each charge state is presented in Table~\ref{train_sample}. %Each sample is consists of 5000 datapoints with 
Input resolution of our dataset is up to 4 decimal places along $m/z$ axis (whereas DeepIso accepts only  2 digits after the decimal point). For training the IsoGrouping module, we cut sequence of frames (each frame holding an isotopic trace) from these peptide features. The detailed procedure and training results are  discussed later in Method section. We apply $k=3$ fold cross validation~\cite{kuncheva2004combining} technique to evaluate our proposed model.
\begin{table}[ht]
\centering
%\resizebox{\textwidth}{!}{
\begin{tabular}{|c|c|c|c|c|c|c|c|c|c|c|}
\hline
Class (charge state) & 1 & 2 & 3 & 4 & 5 & 6 & 7 & 8 & 9 \\
\hline
Peptide Features & 163,038 & 863,050 & 428,909 & 29,183 & 1,503 & 653 & 179 & 236 & 233\\
\hline
\end{tabular}
%}
\caption{\label{train_sample}Class distribution of peptide features in our dataset consisting of 57 LC-MS maps. }
\end{table}

%First, we train the IsoDetecting module that tries to maximize the class sensitivity on validation dataset. Here the class sensitivity is the percentage of datapoints segmented correctly from each class, where classes belong to charge states $z=0$ to 9. The charge state $z=0$ represents the absence of features. The sensitivity of this class indicates how well the model distinguishes actual features from noisy traces and separates the closely residing features as well. In the second step, the sensitivity of IsoGrouping module is defined as the percentage of features reported with correct number of isotopes. We accept a feature if the monoisotope along with high intensity isotopes are reported correctly. We choose the state of IsoGrouping module that maximizes the percentage of feature-matched MS/MS identifications on validation dataset. In the Method section, we elaborate each of the steps in detail along with the training procedure. Finally, we use the trained IsoDetecting module to scan the test LC-MS maps and pass the resultant potential groups of isotopes to the trained IsoGrouping module, which produces the final list of detected peptide features.

\subsection*{Performance Evaluation of PointIso}
%Up to three levels of \textbf{subheading} are permitted. Subheadings should not be numbered.
We run MASCOT 2.5.1 to generate the list of MS/MS identified peptides and the identifications with peptide score > 25 (ranges approximately from 0.01 to 150) are considered as high confidence identifications~\cite{aoshima2014simple}. For performance evaluation, we compare the percentage of high confidence MS/MS peptide identifications matched with the peptide feature list produced by our algorithm and some other popular algorithms. Since the identified peptides must exist in LC-MS maps, therefore, the more we detect features corresponding to them, the better the performance~\cite{aoshima2014simple, tautenhahn2008highly, sturm2008openms, rost2016openms}. 
%In this testing phase we first scan the LC-MS map by IsoDetecting module. Then another run of scan by IsoGrouping module goes through the potential patterns detected in the first step, and reports the final list of peptide feature1123s. 
The other tools used for comparison are MaxQuant 1.6.17.0~\cite{cox2008maxquant}, OpenMS 2.4.0~\cite{rost2016openms}, Dinosaur 1.2.0~\cite{teleman2016dinosaur}, and PEAKS Studio X~\cite{ma2003peaks}. %(although some of these softwares released newer version by the time we finished this work, feature detection algorithm was not upgraded according to the  developer's note, and few repeated experiments did not show any difference in the result. Therefore we did not repeat the full experiment with newer versions). 
The benchmark dataset is prepared by  Chawade et al.~\cite{chawade2014data}, and we use the MaxQuant parameters published by them. For Dinosaur, default parameters mentioned at their github repository (https://github.com/fickludd/dinosaur)
are used. For OpenMS, we use the python binding pyOpenMS~\cite{rost2014pyopenms, rost2016openms} and follow the centroided technique explained in the documentation. %(https://pyopenms.readthedocs.io/en/latest/feature\_detection.html). 
For all of the feature detection algorithms, we set the range of charge state 1 to 9 (or the maximum charge supported by the tool). %Then the produced feature lists are matched with the %high confidence MS/MS identifications and higher percentage of matching implies better performance. 
%In our experiments with the first version of DeepIso, an identified peptide is considered detected if at least one of those peptide features are detected. In other words, multiple MS/MS identifications having the same sequence were considered as the same entity. However, in our current experiments

\begin{figure}[h!]
\centering
\includegraphics[scale = .57]{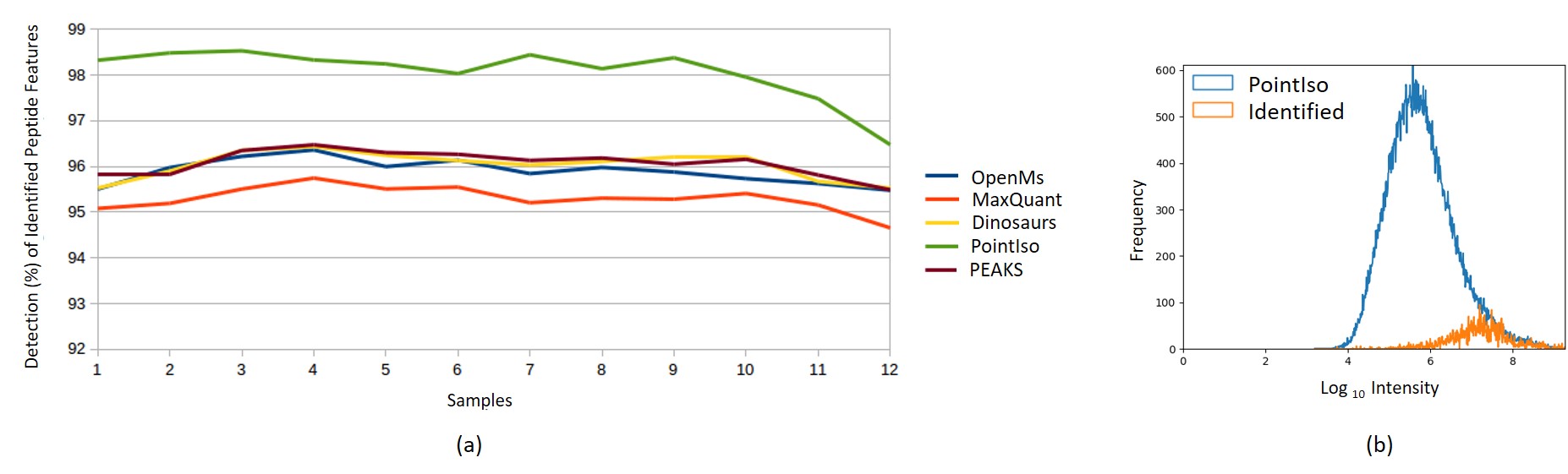}
\caption{(a) Detection percentage of identified peptide features by different tools for 12 samples, each having different concentration. (b) Intensity distribution of peptide features detected by PointIso (blue) and identified  by database search (orange). We see that the distribution of identified features is wedged into the high intensity tail of the distribution of detected peptide features, because only high intensity features are selected for fragmentation. }\label{plot1}
\end{figure}

\subsubsection*{Percentage of Identified Peptide Features Detected by PointIso}

We show the plot of detection percentage of high confidence MS/MS identifications with error tolerance of 0.01 $m/z$ and 0.2 minute peak RT~\cite{zohora2019deepiso,teleman2016dinosaur,chawade2014data} for 12 samples by different algorithms in Figure~\ref{plot1} (a). We see that the PointIso has significantly higher detection rate for all the samples and on average 98.01\%
%99.13\% 
as presented in the first row of  Table~\ref{test_avg} (entire result can be found in Supplementary Table S3). If we just match all the MS/MS identifications (any score) with the peptide features, our algorithm consistently provides higher detection rate than other tools as presented in the second row. Multiple MS/MS fragments coming from  different peptide features can be matched with the same peptide sequence during the database search. Unlike the experiments with DeepIso, here we treat those MS/MS identifications as different entities if their  $m/z$ and peak RT values are significantly different, as it seems more appropriate. %Finally, we tighten the error tolerance by considering 0.005 $m/z$ tolerance along the $m/z$ axis, and our PointIso algorithm achieves consistently higher detection rate as before, as presented in the third row. 
%Next, we take the top 20\% higher abundance peptide features and perform the matching since higher abundance features are important to detect for industrial purpose. Again PointIso is providing higher detection percentage as presented in the third row. 
 We further want to emphasize the fact that, although the model is trained on sample features from certain concentration (e.g., sample 5, 6, 7, 8), but it can detect features having higher or lower concentration as well (e.g., sample 1 to 4, and 9 to 12). It implies that the model can well generalize the peptide feature properties irrespective of peptide intensities seen during training time.

\begin{table}[h!]
\centering
%\resizebox{\textwidth}{!}{
\begin{tabular}{|p{50mm}|c|c|c|c|c|c|}
\hline
Matching Criteria & MaxQuant & OpenMS & Dinosaur & Peaks & DeepIso & \textbf{PointIso}\\ 
\hline
MS/MS identifications with high confidence score & 95.24\% & 95.86\% & 96.01\% & 95.66\% & 96.05\%  & \textbf{98.01}\% \\ \hline
All MS/MS identifications  & 93.73\% & 94.03\% & 94.90\% & 94.82\%  & 94.10\% & \textbf{96.98}\%\\
\hline
%All MS/MS identifications with tighten $m/z$ tolerance for matching & 93.21\% & 92.95\% & 94.23\% & 94.38\%  & Not Applicable & \textbf{94.91}\%\\ \hline
\end{tabular}
%}
\caption{\label{test_avg}Percentage of MS/MS identifications matched by feature list produced by different algorithms.  %Since the DeepIso V1 accepts input data with up to 2 decimal places, the third criteria: tighten $m/z$ tolerance of 0.005 $m/z$ is not applicable for DeepIso V1.
}
\end{table}

\begin{figure}[h!]
\centering
\includegraphics[scale = .6]{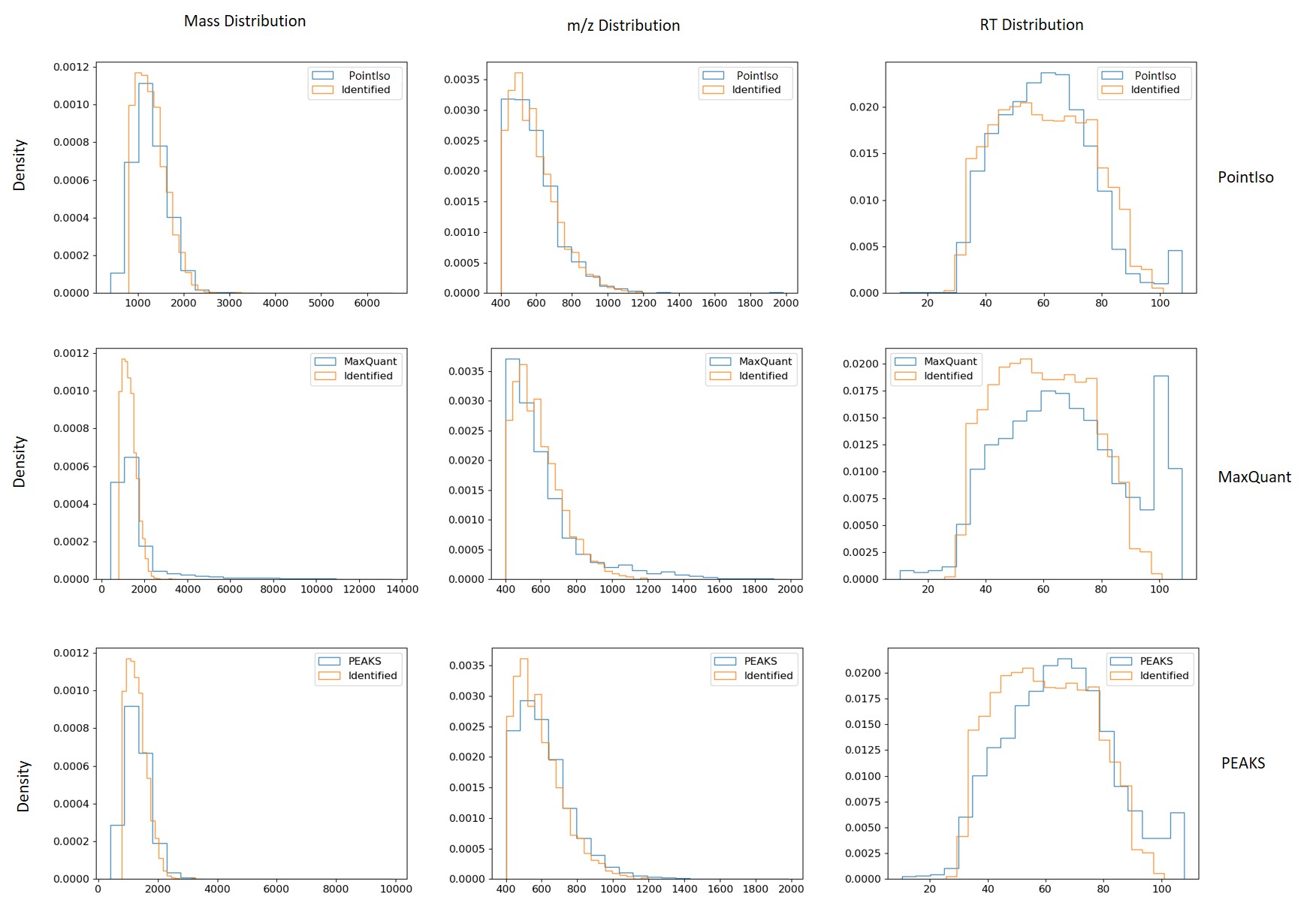}
\caption{Comparison of mass, $m/z$ and RT distribution of detected features (blue) and identified features (orange) for different tools. First, second, and third rows of plots are representing the result for PointIso, MaxQuant, and PEAKS respectively (distributions for other tools are included in Supplementary Figure S2). The orange distributions remain same along the column since its representing the identified peptides. Plots in the first column shows that all algorithms might have some false positives below 1000 Da mass, but the rate is lowest for PointIso. MaxQuant may have some false positives above 2000 Da as well. Then the second column is representing the distribution of $m/z$. Again, MaxQuant might have some false positives in the higher range of $m/z$. Finally, the third column is presenting the distribution of RT and all the tools have a probability of detecting false positives above 100 minute RT, but that rate is lowest for PointIso. Besides that, both MaxQuant and PEAKS might report some false positives below 30 minute RT. For PointIso, the histograms of detected features (blue) are showing good alignment with the histograms of identified features (orange), and implies strong evidence for being true features }\label{distribution_brief}
\end{figure}

\subsubsection*{Quality of Peptide Features Detected by PointIso}
Next we discuss our observation on the peptide features detected by PointIso, but non-identified by database search. The DDA mode ensures that the most abundant peptides are fragmented and thus only those are  tempted for identification. However, the real number of peptide species eluted in a single LC run can be over 100,000~\cite{michalski2011more}. During biomarker identification through relative protein quantification, it is desirable to asses the identity of those remaining peptide features which do not readily match with any MS/MS identification but shows significant abundance change among multiple samples. This is done by performing post-annotation  of  the remaining  peptide features through targeted MS/MS~\cite{jaffe2006pepper}. That is why we wanted to verify whether the PointIso detected but non-identified features are potential peptide features or not. Please note that the typical statistical methods for finding the false positive rate is not applicable here due to the absence of ground truth about the existence of those non-identified peptide features~\cite{tautenhahn2008highly,zohora2019deepiso}. However, one reliable technique is to observe their intensity distribution which should be log normal and if there are many false positives then there will be multiple peaks in the distribution~\cite{michalski2011more}. We present the intensity distribution of the detected features by PointIso in an LC-MS/MS run in Figure~\ref{plot1}(b) which is quite well behaved. %We see that the distribution of identified features (orrange) is wedged into the high intensity tail of the  distribution of detected peptide features (blue). Because only high intensity features are selected for fragmentation as said before. 
Besides this, we also investigated the physicochemical properties of the two populations (blue and orange) as shown in Figure~\ref{distribution_brief}. It demonstrate that the PointIso detected peptide-like features indeed represent peptides~\cite{michalski2011more}. Therefore, we state that PointIso has low false positive rate besides having high detection rate of identified peptides. %Later in Discussion section, we will also explain how to control  false positive rate by fine tuning our deep learning model with appropriate training data. 

%We would like to mention that, retraining with misclassified cases promotes better learning in our model. If a deep learning network makes mistakes, we can collect those cases and feed those back as input samples for fine tuning. This improves the model's ability to give the correct result next time. Retraining was applied in both modules and altogether it improves the model's detection ability by about 8\% (elaborated in Discussion section). Therefore, our model should evolve as new cases appear.

\subsubsection*{Peptide Feature Intensity Calculation by PointIso}
Next we would like to verify the correctness of peptide feature intensity calculation by our model. %For the statistical analysis of biological experiments the peptide feature intensity is to be calculated from raw data~\cite{tautenhahn2008highly}. %The technique is to first apply curve fitting over the bell shaped intensity signals of isotopes in a feature. Then the Area Under Curve (AUC) of all isotopes in a feature are calculated and added to get the intensity or AUC of that feature. Therefore 
The perfectness of peptide feature intensity depends on whether the bell shaped signals are detected nicely or not. We report the Pearson correlation coefficient of the peptide feature intensity, calculated as Area Under the Curve (AUC), between PointIso and other existing algorithms in Table~\ref{pep_intensity}. It appears that our algorithm has a good linear correlation with other existing algorithms, which 
implies that the peptides which are measured differentially among different samples by other tools will also be  reported differentially by PointIso. Thus it validates the peptide feature intensity calculation by our model.
 
\begin{table}[h!]
\centering
%\resizebox{\textwidth}{!}{
\begin{tabular}{|c|c|c|c|c|}
\hline
        & Dinosaur & MaxQuant & OpenMS & PEAKS \\ 
\hline
PointIso & 89.88\% & 95.31\% & 93.76\%  & 88.93\%\\
\hline
\end{tabular}
%}
\caption{\label{pep_intensity} Pearson correlation coefficient of the peptide feature intensity between PointIso and other tools.}
\end{table}

%We also present intensity distribution of reported features in Figure~\ref{dist}. It shows the sensitivity of different tools in different intensity range. If we wish to detect more features from some specific range, we believe our model can be made to do that just by retraining the model (i.e., fine tuning) with more features having that particular intensity range. 

%\begin{figure}[h!]
%\centering
%\includegraphics[scale = .5]{feature_distribution_comparison.jpg}
%\caption{Intensity distribution (AUC) of peptide features reported by different algorithms. }\label{dist}
%\end{figure}

\subsubsection*{Time Requirement of PointIso}
Total time of scanning LC-MS map by IsoDetecting module and IsoGrouping module is considered as the running time of PointIso model. The running time of different algorithms along with the platforms used in our experiment is presented in Table~\ref{runtime}. It appears that our PointIso model is about three times faster than DeepIso and has a comparable running time with most of the existing tools. Although PEAKS is much time efficient than all other tools, we believe PointIso can be made faster by using multiple powerful GPU machines in parallel.  
%give the table
\begin{table}[ht]
\centering
%\resizebox{\textwidth}{!}{
\begin{tabular}{|c|c|c|c|c|c|c|}
\hline
Platform & \multicolumn{3}{c|}{\thead{Processor: Intel Core i7, 4 cores \\OS: Windows 10 for running the applications}}  & \multicolumn{3}{c|}{\thead{Processor: Intel(R) Xeon(R) Gold 6134 CPU, NVIDIA Tesla\\OS: Ubuntu 16.04.5 LTS for running the python scripts}} \\
\hline
Algorithms &  PEAKS & Dinosaur & MaxQuant  & \textbf{PointIso} & DeepIso & OpenMS \\
\hline
Running Time & 8 minutes & 15 minutes & 30 minutes & \textbf{30 minutes} & 1 hour and 40 minutes & 2 hours and 50 minutes  \\
\hline
\end{tabular}
%}
\caption{\label{runtime}Approximated running time of different algorithms. Here the platform used for OpenMS, DeepIso and PointIso did not have support for running Windows application of PEAKS, MaxQuant, and Dinosaur. So we used different machine for running those.}
\end{table}

Finally, we  want to emphasize the fact that, our PointIso is able to learn the general peptide feature properties irrespective of the concentration of protein samples provided during the training. Because the twelve LC-MS maps in the dataset comes from three different range of protein concentration: low, medium, and high. We train over one range and run the test on two other ranges. The average detection results are already discussed and we see that the PointIso consistently provides higher detection rate. It implies that, once we train a model on a protein sample, the same model should be applicable to all other protein samples from the same or other close species without further training. This should make PointIso more appealing in the  practical sectors. 

%We present a table showing complete breakdown of different developmental stages and referred in our discussion. 
\section*{Discussion}
%The Discussion should be succinct and must not contain subheadings.
We propose PointIso, a deep learning based model that discovers the important characteristics of peptide feature by proper training on vast amount of available LC-MS data. Other heuristic algorithms have to set different parameters, e.g., number of scans to be considered as feature, centroiding parameters, theoretical formulas for grouping together the isotopes, and also data dependant parameters for noise removal and other preprocessing steps. On the other hand, PointIso does not rely on manual input of these parameters anymore and systematically learns all the necessary parameters itself. We will first demonstrate the justification of different design strategies performed. Then we will discuss some prospective research directions. We show a complete breakdown of different developmental stages in Table 5, which is referred in different sections of the following discussion.   
%We present a table showing complete breakdown of different developmental stages and referred in our discussion. 
%give the table
\begin{table}[h!]
\centering
%\resizebox{\textwidth}{!}{
\begin{tabular}{|c|c|c|}
\hline
Model & Matching with MS/MS identified peptides \\
\hline
Initial model & 65\% \\
\hline
Bi-directional 2D RNN & 72\% \\
\hline
Dual attention mechanism  & 94\% \\
\hline
Retraining of IsoDetecting module with long RT range & 95.5 \% \\
\hline
Increasing resolution from 0.01 $m/z$ to 0.0001 $m/z$ & 97\% \\
\hline
Retraining using features detected with wrong charge by IsoDetecting module & 98.22\% \\
\hline
New architecture of IsoGrouping module & 99.55\% \\
\hline
Fine tuning with feature like noises & 98.52\% \\
\hline
\end{tabular}
%}
\caption{\label{progress}Performance of PointIso in different developmental stages (based on validation dataset).}
\end{table}

\subsection*{Point Cloud Representation of IsoDetecting Module with Weighted Cross Entropy Loss}
%One big concern to move from image representation (first version) to point cloud  representation of the datapoints was to achieve the capability of processing high resolution data in a timely manner.
In DeepIso, IsoDetecting module process the 2D image representation of LC-MS map by sliding a window pixel by pixel, and producing a output $z=0$ to 9 at each step using classification network, depending on whether there is a feature aligned with the window or not. Accepting higher resolution input is feasible if we  formulate IsoDetecting network as a segmentation network so that one scanning window can predict all datapoints in it at a time. Now, a scanning window covers 15 RT scans and 2.0 $m/z$ having resolution of up to four decimal points. %Therefore, 2D image representation with 0.01 $m/z$ resolution will have $15 \times \frac{2.0}{0.01} = 3000$ pixels to segment. Since IsoDetecting module must be made compatible to accept higher resolution input (0.0001 $m/z$) for improving the feature detection, total number of pixels in a scanning window becomes $15 \times \frac{2.0}{0.0001} = 300,000$. However, actual number of datapoints in a scanning window is about 5000 , even with that high resolution. 
With point cloud representation it has to predict the label of about 5000 points only (the $95^{th}$ percentile of the number of datapoints in all the sample scanning windows), whereas it needs to predict 300,000 pixels with 2D image representation ($15 \times \frac{2.0}{0.0001} = 300,000$), where most of the points will be blank.  So image representation makes the segmentation problem unnecessarily voluminous.  Therefore to support higher resolution compatibility at a faster speed, we move from image to point cloud representation of datapoints. There are also other literature, e.g., DeepNovoV2~\cite{qiao2019deepnovov2}, which switched to point cloud representation for supporting higher resolution data like us.

%\subsection*{Weighted Cross Entropy Loss in IsoDetecting Module}
Unlike DeepIso, we had to deal with highly class-imbalanced problem with this new representation of IsoDetecting module.  
Only 20\% of 5000 input datapoints are positive and all other datapoints are noisy signals or just have  zero intensity. However, a positive datapoint might also be recorded as zero or low intensity point due to instrumental noise. So we cannot just disregard those datapoints. %Because sometimes a isotopic signal is not recorded due to instrumental noise and although it appears zero (or very low), but we still have to scan through it and in the output it is supposed to be predicted as positive point. Since during the testing stage, we will not know in advance which zero intensity points may belong to such positive class, we cannot just discard those to increase the positive class ratio. However, we do not know in advance which zero intensity points are actually background point and which  %to prevent the unnecessary break in the corresponding feature. 
%Therefore during the training, we actually label those 0 intensity signals as positive charge $z$, so that model learns to detect it as a single continuous peptide feature instead of breaking it into two separate peptide features. 
Therefore, we use class weights while calculating cross entropy loss so that both the positive and negative datapoints are learnt well. We use class weights decided based on the distribution per sample, since it tries to achieve well balanced class sensitivity according to our experiments (included in Supplementary Note A).  %Therefore we choose this class weighting approach throughout our work. 
This initial model was able to achieve  about 65\% matching with the peptide identifications as shown in Table~\ref{progress}. Therefore we were in need of further investigations for the improvement which are discussed in next section.

\begin{figure}[h!]
\centering
\includegraphics[scale = .6]{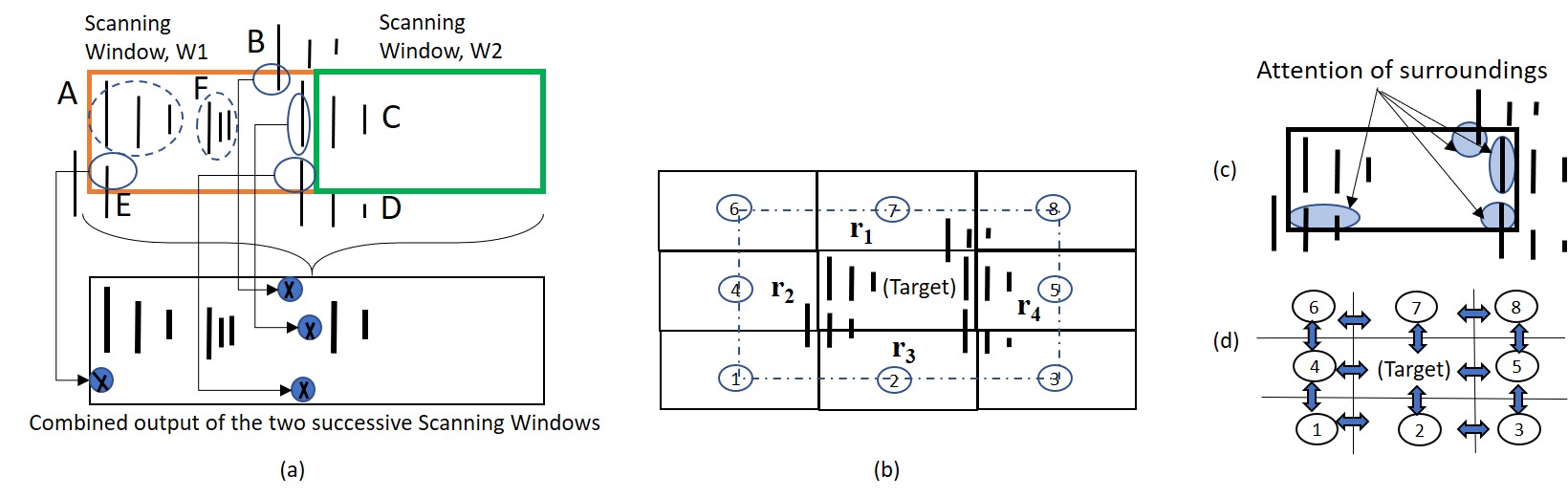}
\caption{(a) We show two successive and non-overlapping scanning windows, W1 and W2, and six features: A, B, C, D, E, and F. We show the combined output of the two successive scanning windows in the bottom image. Feature A and F are fully contained within W1. Therefore, all of it’s isotopic signals are correctly detected, as shown in the combined output. However, for each of other four features, W1 sees partial traces shown by small circles in the upper image. Without any background knowledge, those traces are not adequate for deciding whether they belong to real features or merely noisy traces. We mark the isotopes by cross sign which are not detected. The second window W2 detects two isotopes of feature C. But the system missed the monoisotope of feature C, which is treated as missing the feature as a whole. (b) Surrounding regions of a target window. (c) Attention of surrounding regions over the datapoints of target window. (d) 2D bi-directional RNN to flow the surrounding information towards the target window in center.}\label{atntn}
\end{figure}
%table related to model core architecture 

% table related to weighting
\subsection*{Attention Mechanism}
\begin{figure}[h!]
\centering
\includegraphics[scale = .6]{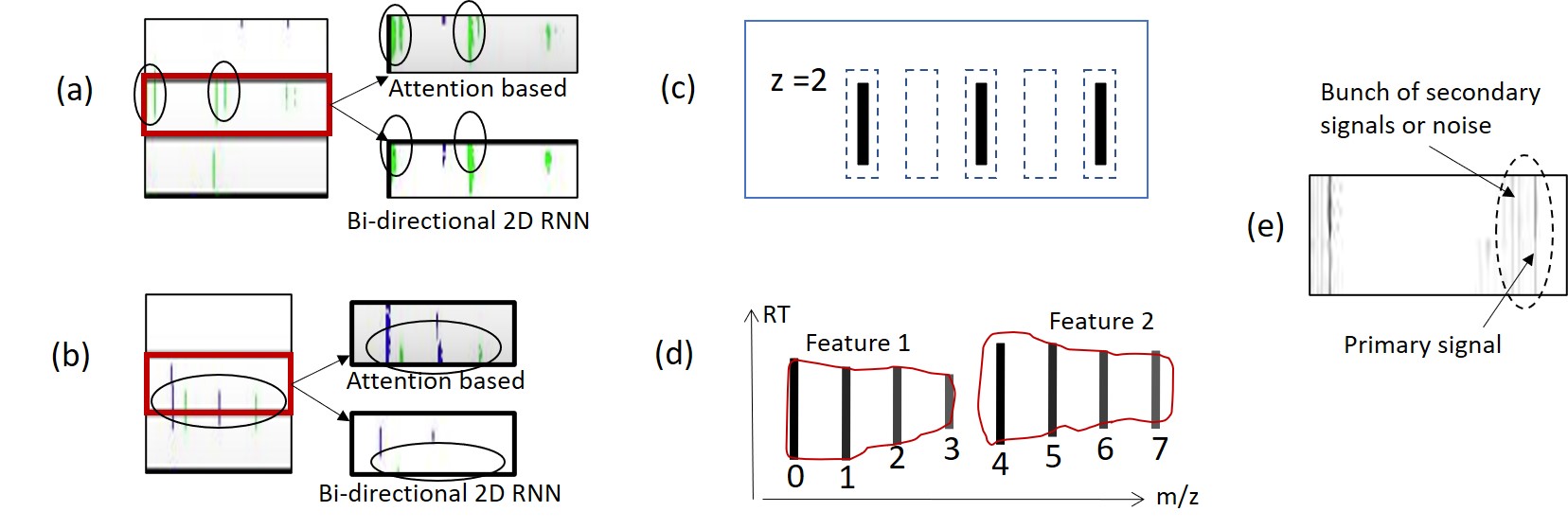}
\caption{(a) and (b) are showing the comparison between attention based mechanism and bi-directional 2D RNN. The Red rectangle is showing the target window. Detection by attention mechanism and bi-directional two dimensional RNN for each target window are shown next to it, pointed by arrow signs. We see that attention mechanism works better in separating closely residing features in (a), and detecting partially seen features in (b). (c) When isotope lists are passed to the IsoGrouping module with wrong frames (dotted rectangles) due to the wrong charge ($z=4$) detected by IsoDetecting step, it results in discarding this whole group of frames as a  noise due to the inconsistency (blank frames) observed. (d)~Adjacent feature problem.  (e) Feature like noisy signals }\label{fine}
\end{figure}

Next, we discuss the reason of using attention mechanism with segmentation network of IsoDetecting module. In a random scenario feature can spread over multiple windows, and simply using segmentation network without any surrounding knowledge will cause missing of the features, e.g., feature C, as illustrated in Figure~\ref{atntn}(a). It also fails to compute total abundance of features perfectly, since its missing trailing regions of features like B, D, and E. %We already mentioned that the best case scenario during scanning occurs  when the feature is aligned with the left boundary of window, e.g., feature A. The network also produce 98\% positive class sensitivity in that case, as shown in the last row of Table~\ref{weight}.  However, 
To overcome this problem we have to incorporate surrounding knowledge while segmenting the datapoints of a target window, i.e., W1 in Figure~\ref{atntn}(a). 
\begin{table}[ht]
\centering
%\resizebox{\textwidth}{!}{
\begin{tabular}{|c|c|c|c|c|c|c|}
\hline
Experiment Category & z=0 & z=1 & z=2 & z=3 & z=4 \\
\hline
Sliding window with 50\% overlapping & 86\% & 50\% & 72\% & 57\% & 36\%  \\
\hline
Skiplink inserted in above model & 85\% & 56\% & 76\% & 66\% & 41\% \\
\hline
Bi-directional 2D RNN & 85\% & 51\% & 77\% & 67\% & 49\% \\
\hline
\textbf{Attention mechanism} & 84\% & 62\% & 85\% & 71\% & 50\% \\
\hline
\textbf{Attention mechanism with higher resolution} & 90\% & 64\% & 85\% & 81\% & 61\% \\
\hline
\end{tabular}
%}
\caption{\label{atn_table} Different techniques of absorbing surrounding information and corresponding class sensitivity of IsoDetecting module. We define the class sensitivity of a scanning window as the number of datapoints from class $z$ (0 to 9) detected correctly out of total number of datapoints in a scanning window. To evaluate candidate solutions we use
the class sensitivity of high abundant features (charge $z=$ 1,2,3, and 4) in a 
\emph{average case} scenario. Average case means the scanning window might contain any number of features, they may appear at any location of the window, they might be partially or fully seen, and might be overlapping as well. We see that the DANet inspired attention based mechanism works better than other techniques.}
\end{table}
According to our experiments, we find that the regions $r_1$, $r_2$, $r_3$, and $r_4$ in Figure~\ref{atntn} (b) are actually playing the key role in detecting the traces inside target window. Our experimental results with three different criteria: 50\% overlapping of scanning window, Bi-directional 2D RNN (Figure~\ref{atntn} (d)), and the attention mechanism (Figure~\ref{atntn} (c)) inspired by DANet are presented in Table~\ref{atn_table}. %(details of experimental procedure and result are provided in supplementary materials).
 Besides that, we also had a visual verification about whether the partially seen peptide features are properly detected or not as presented in Figure~\ref{fine} (a) and (b). Since PointNet segmentation network combined with DANet works better than other techniques, we choose this to develop our IsoDetecting module.

%table related to model core architecture 

\subsection*{Upgrading IsoGrouping Module}
%We bring some changes in the IsoGrouping module to support the higher resolution output from IsoDetecting module. 
IsoDetecting  module outputs the isotope list with $m/z$ resolution of up to 4 decimal places, but IsoGrouping module does not need that much high resolution to group together the potential isotopes into a feature. Therefore, we use a resolution degrading approach before passing the isotope lists to IsoGrouping module. But we ensure that the isotopes who merge together in lower resolution are kept in separate list and thus passed to IsoGrouping module separately. This resolves the problem of missing features merged in lower resolution, one of the key problems of DeepIso. Besides that, in that model each frame of the input sequence covers a wide range of $m/z$ value. As a result, each frame holds an isotopic signal along with it's background. %  such that it covers the range (peak intensity point - 0.04) to (peak intensity point + 0.04) along $m/z$ axis and 15 RT scans. Therefore, the frame also covers the background noise besides the signal. 
But in PointIso, we filter out the signal area from background, based on the boundary information provided by IsoDetecting module. %by taking (peak intensity point - 10 ppm) to (peak intensity point + 10 ppm) range along $m/z$ axis. 
So the background of input frames are mostly blank, and IsoGrouping module can see the bell curve better than before. As a result we can get rid of the attention gate in this module to keep it simpler. Another change is, instead of using RNN layer to process frames of the sequence one at a time, we process five frames at a time using a network consisting of CNN and fully connected layers through weight sharing. We observe that it results in better prediction at the output layer. Besides that, we incorporate Area Under the Curve (AUC) of isotopic signal as context information through embedding which reduces the uncertainty during class prediction. The charge, detected in previous step is fed into the network through a scaling gate (neuron) which helps in proper grouping as well. The final architecture of IsoGrouping module is quite different and presented later in Method section. It  improves the feature detection by about 1.5\%, as presented in the seventh row of Table~\ref{progress}.
%The output layers from five frames are concatenated and three more layers of fully connected layers are used after that to produce the final output. Thus its able to find the frame that holds the ending isotope of the feature. If feature is consisting of more than five isotopes, we slide the window further to find those. Details are explained in Method section.  

\subsection*{Fine Tuning Using Misclassified Features}
Retraining the primary model by feeding back the misclassified data played an essential role in overall improvement. We applied this approach in learning four particular cases. First, some features having very long RT range were not detected or broken into multiple features by IsoDetecting module. Therefore, fine tuning the model with such samples improves the detection rate by about 2\% (fourth row in Table~\ref{progress}). Second, 
IsoDetecting module was predicting wrong charge for the features like `F' in Figure~\ref{atntn} (a), which appears in the middle region of the scanning window, i.e., not aligned with the scanning window. As a result, wrong set of frames are passed to the IsoGrouping module, as shown in Figure~\ref{fine} (c). This causes complete rejection of the feature by IsoGrouping module, and we miss a feature although it was detected in the first module.  %(might or might not fully contained within the scanning window)
 %Some of the features like that were detected with a wrong charge due to the increased uncertainty introduced by segmentation network. 
 %, because ideally each frame is supposed to hold an isotopic trace. 
So, retraining the IsoDetecting module using such samples improves the detection rate by about 1.5\% (sixth row in Table~\ref{progress}). Third, adjacent features as shown in Figure~\ref{fine}(d) were not correctly separated  by IsoGrouping module.  Here, both features have same charge, same or very close RT value at peak intensity point, and the distance between the two features along $m/z$ axis is equal to the distance between their own isotopes. Fine tuning IsoGrouping module with such samples resolves this problem.
%was actually introduced in version one, and we do it in this version from the very beginning. 
Finally, we fine tune the model with feature like noisy or secondary signals (e.g.,  Figure~\ref{fine} (e)) which appear very close to the main isotopic signal. Although PointIso reports 99.55\% detection of  identified peptides without this fine tuning, but it also reports multiple features for basically the same peptide feature. %It results in about 200,000 total peptide features and 99.55\% detection of  identified peptides. 
After training with such samples the problem is solved with % total number of peptide features  reduces to about 100,000 with 98.52\% detection of identified peptides.
98.52\% detection of identified peptides (discussed further in the Supplementary Note C). It should be an interesting research scope to see if we can solve this problem without reduction in detection percentage. Besides that, we believe the necessity of this step also depends on the dataset and the mass spectrometer used. Because high resolution mass spectrometers usually produce narrower signals without those secondary peaks,  thus we might avoid this step and obtain over 99\% detection. %We can actually balance between true positive rate and false positive rate according to our expectation through fine tuning using appropriate amount of training samples (discussed further in the Supplementary Note C). This should be an interesting research scope to see if we can achieve low false positive rate without this reduction in the detection percentage. 
\begin{figure}[h!]
\centering
\includegraphics[scale = .4]{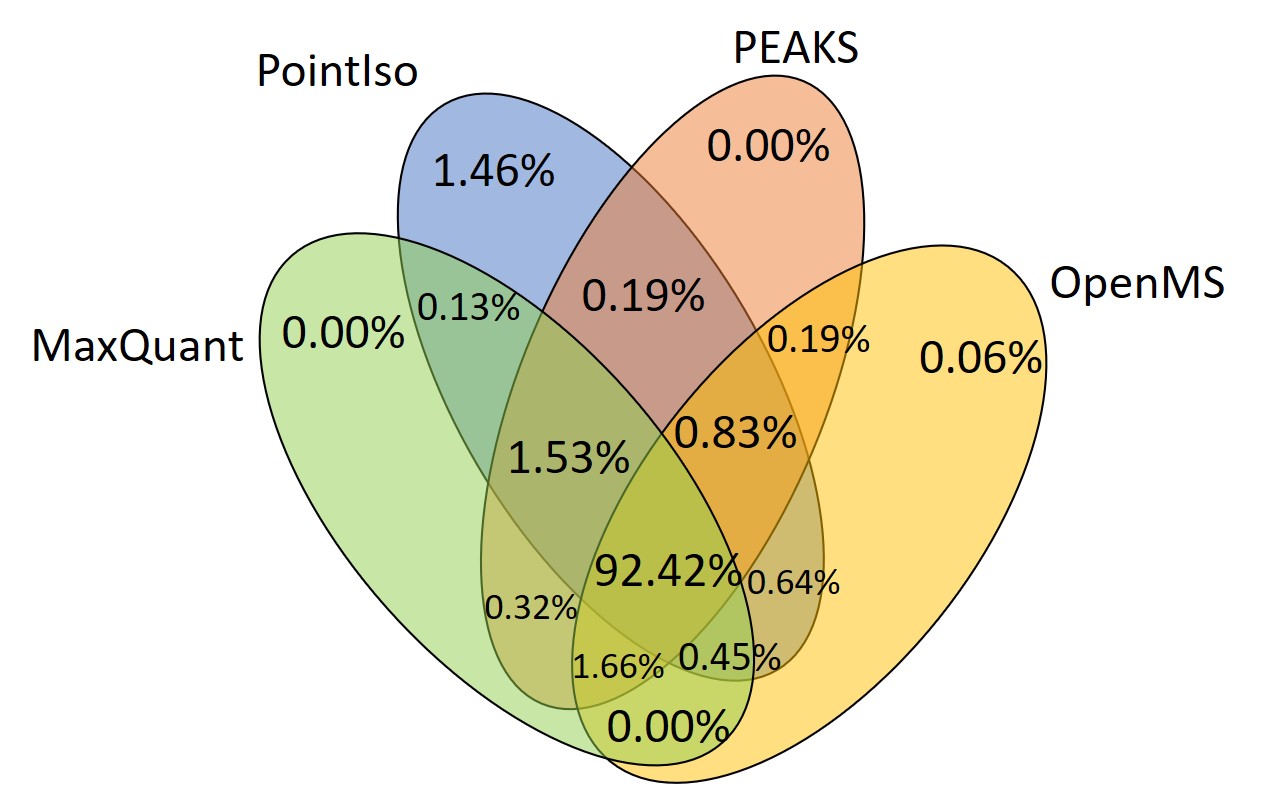}
\caption{Venn diagram of identified peptide features detected by different algorithms. }\label{venn}
\end{figure}
\begin{figure}[h!]
\centering
\includegraphics[scale = .5]{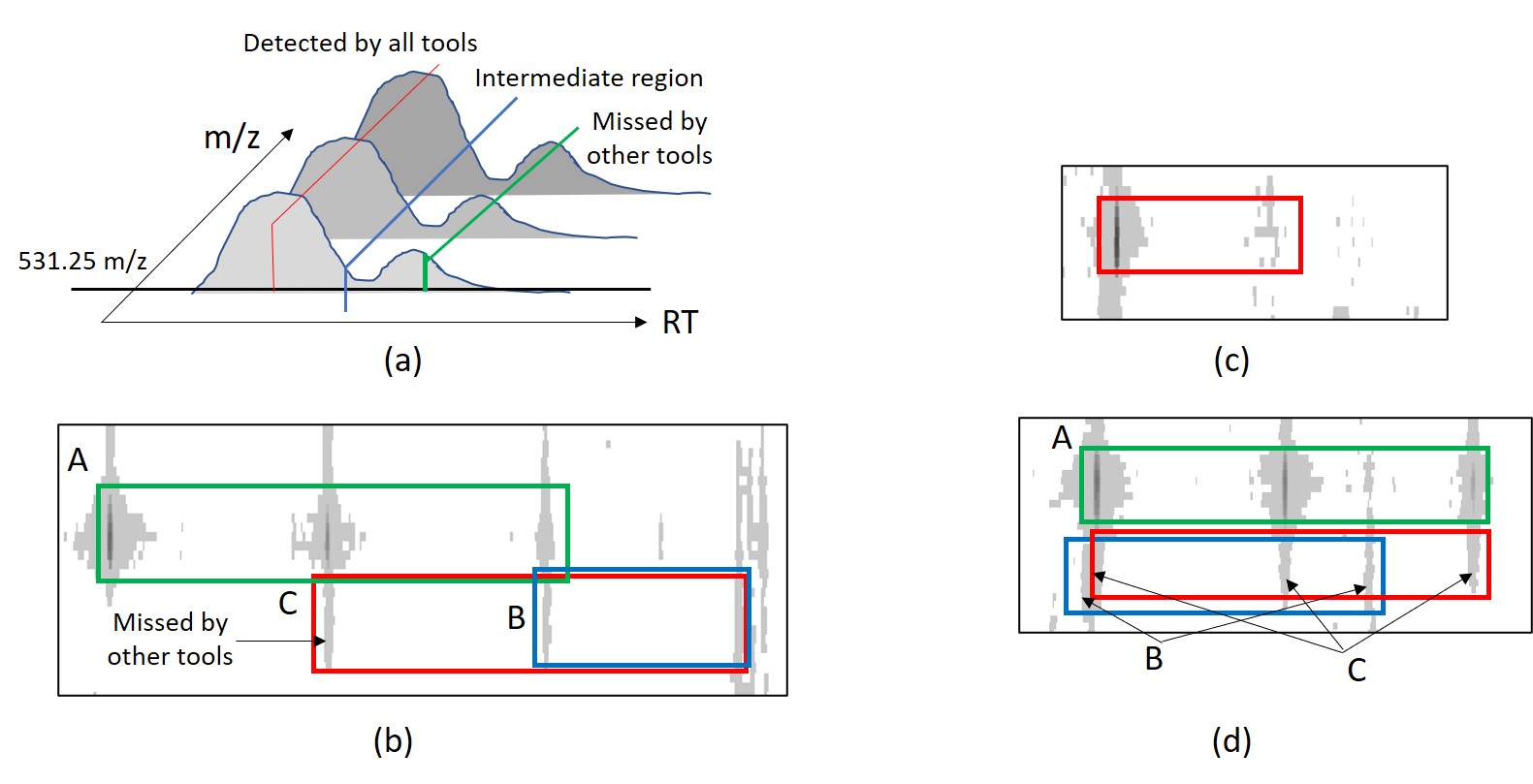}
\caption{Observations on peptide features detected by only PointIso. (a) Peaks connected by red line are detected as a peptide feature by all algorithms. However, the feature having lower peak, connected by green line, is missed by other tools. Because they have the same $m/z$ as the one with higher peak. Therefore, most of the tools merge it with the bigger one during pre-processing steps. (b) Monoisotope of the feature enclosed in red rectangle is missed by other tools due to merging with A. But PointIso detected both feature A and C. (c) Features with broken signals are detected by our model. (d) Very closely residing and overlapping features, like feature in blue and red rectangles are sometimes missed by other tools as well, although detected by PointIso. }\label{venn_img}
\end{figure}

\subsection*{Observations on Peptide Features Exclusively Detected by PointIso}

We present the Venn diagram of identified peptide features detected by different algorithms in Figure~\ref{venn} (we show four algorithms to keep the Venn diagram simple). We visually inspected all the features which are detected by PointIso only. We found that the features whose peaks are very close to each other along RT axis, are sometimes merged into one in other tools, e.g., Figure~\ref{venn_img} (a), but separated perfectly by PointIso. Similarly, the monoisotope might be missed sometimes by other tools as shown in Figure~\ref{venn_img}(b). Here, the feature A and C are detected by PointIso. But other tools detect A, and instead of C they report B by mistake, missing the monoisotope (merged with A). Very closely residing and overlapping features, like feature C in Figure~\ref{venn_img} (d) are sometimes missed by other tools as well, although detected by PointIso. Besides these, PointIso can detect features even if the signals  are not recorded perfectly as Figure~\ref{venn_img}(c), which are sometimes discarded by other algorithms.   
%Although batch normalizing is a popular technique for achieving faster convergence and also used in the original paper, however, in our case the data distribution varies drastically between the training and testing case. So batch normalizing parameters, e.g., mean and variance cannot be learned properly and cause drastic fluctuations in validation loss. It prevents us from choosing the best model and thus we 
%don't use batch normalizing layers. our training takes about 80 epochs to converge with about 300k training data. Batch normalization did not worked.
\\
\\
Finally, we refer to the fact that, although we have prepared the training data by taking the common set of Dinosaur and MaxQuant in order to replace human annotators, but the outcome of PointIso is quite different than those algorithms. Because deep learning network learns the required parameters by stochastic gradient descent through several layers of neurons and backpropagating the prediction errors, a completely different technique than all the existing heuristic methods. That is why PointIso achieves higher detection rate than others. Some appealing future works should involve plugging in PointIso in the downstream workflow of peptide quantification, identification of chimeric spectra, and disease biomaker identification. Extending this model for Intact Mass analysis might also be another important research direction. Also it will be interesting to see how the attention based non-overlapping sliding window approach performs in general object segmentation problems. We are looking forward to these research opportunities in the future.

\newpage
\section*{Methods}
%Topical subheadings are allowed. Authors must ensure that their Methods section includes adequate experimental and characterization data necessary for others in the field to reproduce their work.
Our model runs the processing on raw LC-MS map which is obtained in .ms1 format using the ProteoWizerd 3.0.18171~\cite{chambers2012cross}. Then we read the file and convert it to a pointcloud based hash table where RT scans are used as keys and (m/z, intensity) are inserted in a sorted order under those keys. Therefore we have the datapoints saved as triplets (RT, $m/z$, intensity) in the hash table.

\subsection*{Step 1: Scanning of LC-MS map by IsoDetecting module to detect isotopes}  

%Our job of scanning the LC-MS map along the RT axis resembles the video clip classification, where the RT axis is the time horizon. We build the module combining the CNN and RNN in a FC-RNN fashion proposed by Yang et al.~\cite{yang2016multilayer}, which achieved state-of-the-art results in the context of video classification on two benchmark datasets. 

\begin{figure}[h!]
\centering
\includegraphics[scale = .7]{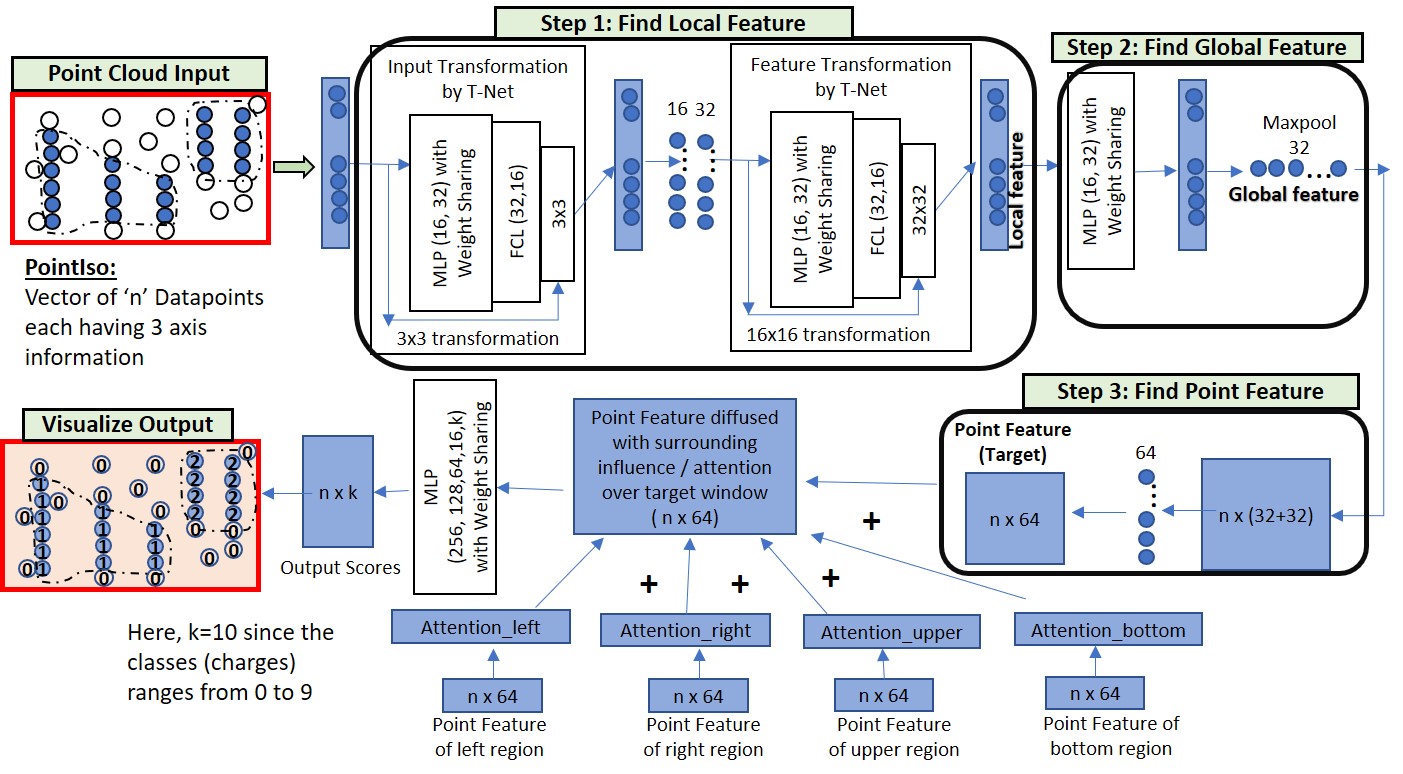}
\caption{Network of IsoDetecting module. This network goes through three steps, finidng  the local features, global features and point features respectively of the given target window. The number of layers and neurons in the Multiple Parceptron Layers (MLP) and Fully Connected Layers (FCL) are determined by experiments and mentioned in the figure. Point features of target window are then diffused with features of surrounding regions based on their attention or influence over target window (calculation of $Attention\_left$ and others are shown in next figure). Finally the diffused features are passed through four MLP and softmax layer at the output provides the final segmentation result.}\label{pointnet}
\end{figure}

\begin{figure}[h!]
\centering
\includegraphics[scale = .5]{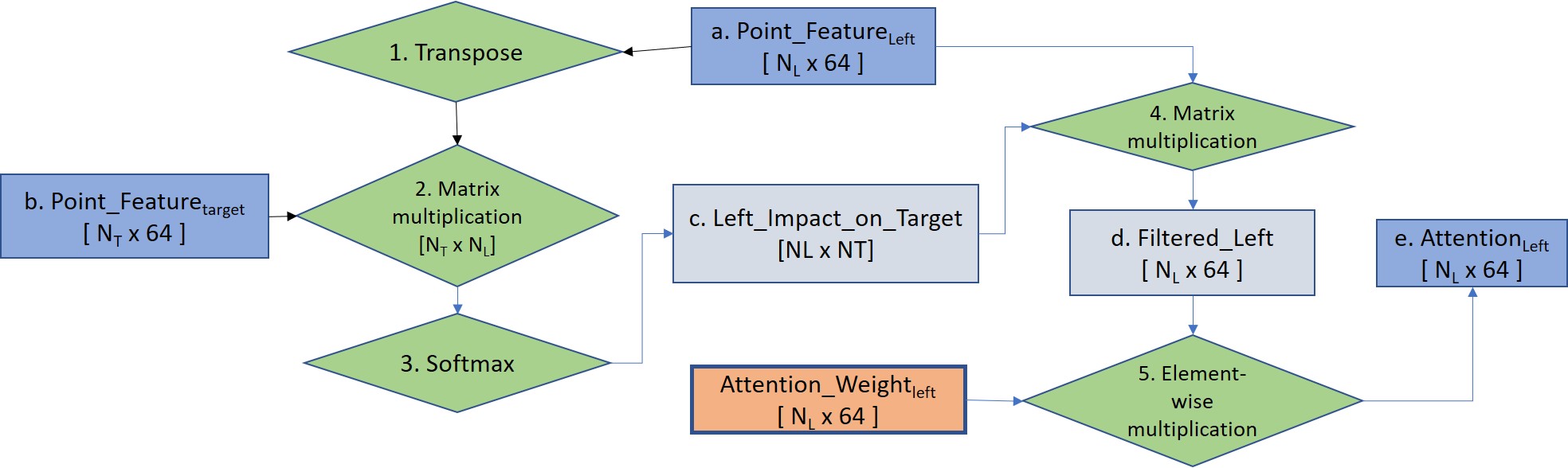}
\caption{Flowchart of attention calculation in IsoDetecting module. This particular flowchart is intended to find out the attention or impact of left region over the datapoints of target window. Exactly similar approach is followed for other surrounding regions as well and finally all are diffused with the $point\_features\_target$ by addition.}\label{danet}
\end{figure}

Our network scans the 3D LC-MS plot using non-overlapping  sliding window having dimension 2.0 $m/z$, 15 RT scan, and covering full intensity range, as already presented in Figure~\ref{alg}. The intensities are real numbers scaled between 0 to 255. Here the objects, i.e., peptide features  are to be separated from background. Background may contain feature like noisy signals and peptide features are frequently overlapped with each other. So the target is to label each datapoint represented by triplet (RT, $m/z$, intensity) with its class. The class is either charge $z=1$ to 9 (positive) if the datapoint belongs to a feature having that charge, or $z=0$ if that datapoint comes from background or noise.  %The map is divided into $N=total mz range / 2.0 $ stripes. We slide a window to scan each stripe in a bottom up order. 
Each window sees a  point cloud which is essentially a set of points, or triplet (RT, $m/z$, intensity). This is passed through a PointNet architecture as shown in Figure~\ref{pointnet}. In order to properly segment peptide features spreading over multiple sliding windows, we adapt the DANet~\cite{fu2019dual} and plug into our model to find attention or influence of four surrounding regions (Figure~\ref{atntn} (b)) over the target window datapoints. We present a flowchart in Figure~\ref{danet}, showing the calculation of attention coming from the surrounding regions. The detailed explanation of this flowchart is provided in Supplementary Method A.% (addition among other diffusion techniques is chosen based on experiment).

We can divide the LC-MS map along $m/z$ axis into sections of equal ranges and process multiple sections in parallel to make the process time efficient. We keep nine hash tables for recording the detection coordinates (RT, $m/z$) of features from nine classes ($z=$ 1 to 9) during the scanning. The $m/z$ values of the isotopes are used as the key of these hash tables, and the RT ranges of the isotopes in a feature are inserted as values under these keys as shown in the block diagram of Figure~\ref{alg}. Since the detection of wider isotopes may span over a range of points along $m/z$ axis, we take their weighted average to select specific $m/z$ of an isotope.

\subsubsection*{Training Procedure of IsoDetecting module}
IsoDetecting module is supposed to learn some basic properties of peptide feature (provided in Supplementary Note D), besides many other hidden characteristics from the training data. Each sample consists of a target window and its four surrounding regions. As already mentioned, we select the common list of peptide features provided by MaxQuant and Dinosaurs, to generate the positive samples. We slide the scanning window over the feature and its surrounding region so that we can generate samples %having the feature appearing in every location of every region. As a result we actually generate samples 
holding peptide features in different locations of the target window, along with its surrounding areas. Similarly we choose some areas containing only noises of different intensities, feature like noisy signals, and completely blank areas as well. We call them negative samples. We see the total number of samples used for training in Table~\ref{train_val_sample_1}. Also note that, unlike DeepIso, IsoDetecting module is segmentation network here. As a result it has to classify each datapoint in a sample. One sample window prepared from feature having charge `2' might contain features with other charges and IsoDetecting module has to classify all those datapoints enclosed by that sample window. Therefore, the class sensitivity is actually impacted by the total amount of datapoints having that class, taking account \emph{all} the training samples. So we have included a column label as `Total Datapoints' in the table.%In this way we generate approximately 150,000 positive samples and 150,000 negative samples. 

%give the table
\begin{table}[ht]
\centering
%\resizebox{\textwidth}{!}{
\begin{tabular}{|c|c|c|c|c|}
\hline
\multirow{2}{*}{Class ($z$)}  & \multicolumn{2}{c|}{\thead{Training}} &  \multicolumn{2}{c|}{\thead{Validation}}\\
\cline{2-5}
 & Total Datapoints & Total Samples & Total Datapoints & Total Samples \\
\hline
0 & 544,014,927 & 40,502 & 22,100,416 & 10,138  \\
\hline
1 & 3,648,512 & 37,924 & 167,671 & 1,282 \\
\hline
2 & 28,633,707 & 152,148 & 1,731,244 & 8,902 \\
\hline
3 & 19,021,011 & 91,542 & 1,033,477 & 4,485 \\
\hline
4 & 2,998,905 & 25,526  & 77,534 & 281 \\
\hline
5 & 3,526,031 & 22,134   & 3,032 & 13 \\
\hline
6 & 431,139 & 7,721 &  606 & 5 \\
\hline
7 & 30,145 & 4,472  & 319 & 4  \\
\hline
8 &  18,144 & 8,706 &  48 & 3  \\
\hline
9 & 17,310 & 3,429 &  227 & 2  \\
\hline
\end{tabular}
%}
\caption{\label{train_val_sample_1} Amount of samples for training and validation. Because of inadequate training data for features with charge states 5 to 9 as presented in Table~\ref{train_sample}, we had to apply data oversampling and augmentation in order to increase training samples from these classes. Amount of samples from class 0 depends on our choice. We chose the amount so that total datapoints from this class is higher than others, because the LC-MS map is very sparse. The validation set does not contain any duplicated data and there is no overlapping between validation dataset and training dataset.
}
\end{table}

The average sensitivity of the trained model on training set and validation set are provided in Table~\ref{train_val_1}. We use minibatch size of 8 during the training, because the network already takes about 15 GB GPU Memory due to the sophisticated architecture. We use `NAdam' stochastic optimization ~\cite{kingma2014adam} with initial learning rate of $0.001$. If we do not observe any significant drop in the validation loss for about 5 epochs, we decrease the learning rate by half. The model converges within about 100 epochs. We use sparse softmax cross entropy as error function at the output layer. Besides that, we apply the class weights as already mentioned in the Discussion section and is elaborated in supplementary material.

%give the table
\begin{table}[ht]
\centering
%\resizebox{\textwidth}{!}{
\begin{tabular}{|c|c|c|c|c|}
\hline
\multirow{2}{8mm}{Class ($z$)}  & \multicolumn{2}{c|}{\thead{Training}} &  \multicolumn{2}{c|}{\thead{Validation}}\\
\cline{2-5}
 & Sensitivity-Average  (\%) & Sensitivity-Best  (\%)  & Sensitivity-Average (\%) & Sensitivity-Best  (\%)   \\
\hline
0 & 88.48 & 30.0 &  88.48 & 38.69 \\
\hline
1 &  57.59 & 91.0 & 61.54 & 99.28  \\
\hline
2 & 76.90 & 97.0 & 82.98 & 98.40\\
\hline
3 &  75.08 & 94.75 &  79.42 & 96.05\\
\hline
4 &  64.78 & 94.24 & 52.80 & 97.33\\
\hline
5 &  88.57 & 94.21  & 58.74 & 100\\
\hline
6  & 60.73 & 82.5 & 15.68 & 70.56\\
\hline
7 &  40.12 & 60 & 10.03 & 70.78 \\
\hline
8  & 4.07 & 10 & 3.0  & 10 \\
\hline
9 & 4.50 & 8 & 2.01 & 15 \\
\hline
\end{tabular}
%}
\caption{\label{train_val_1}Class sensitivity of IsoDetecting module. We show two cases, average and best case in terms of feature detection ability. Best case occurs when the feature is left aligned with the scanning window boundary, e.g., feature `A' in Figure~\ref{atntn} (a). Average case means the scanning window might contain any number of features, they may appear at any location of the window, they might be partially or fully seen, and might be overlapping as well. Due to the lack of variance in training data for charge states 6 to 9, the model's validation sensitivity does not go up high for these classes. However, since most of the peptide features appear with charge states~<~6, lower sensitivity for them does not impact the overall performance. The validation set does not contain any duplicated data and there is no overlapping between validation dataset and training dataset.}
\end{table}

\subsection*{Step 2: Scanning of LC-MS Map by IsoGrouping Module to Report Peptide Feature}

There are four major differences in IsoDetecting and IsoGrouping modules. First, the IsoDetecting module scans the LC-MS map along both the RT and $m/z$ axis, whereas IsoGrouping module scans left to right, i.e., only along the $m/z$ axis. Second, IsoDetecting network is a point cloud based network, whereas the IsoGrouping network is an image based network. Third, IsoGrouping module performs a sequence classification task that generates one output after seeing through 5 consecutive frames, unlike IsoDetecting module which segments the datapoints of the input frame. Last, IsoDetecting module accepts very high resolution $m/z$ values (up to 4 decimal place), but IsoGrouping works on comparatively lower resolution $m/z$ values (up to 2 decimal place), because it does not need such higher resolution to group the isotopes into features.

\begin{figure}[h!]
\centering
\includegraphics[scale = .5]{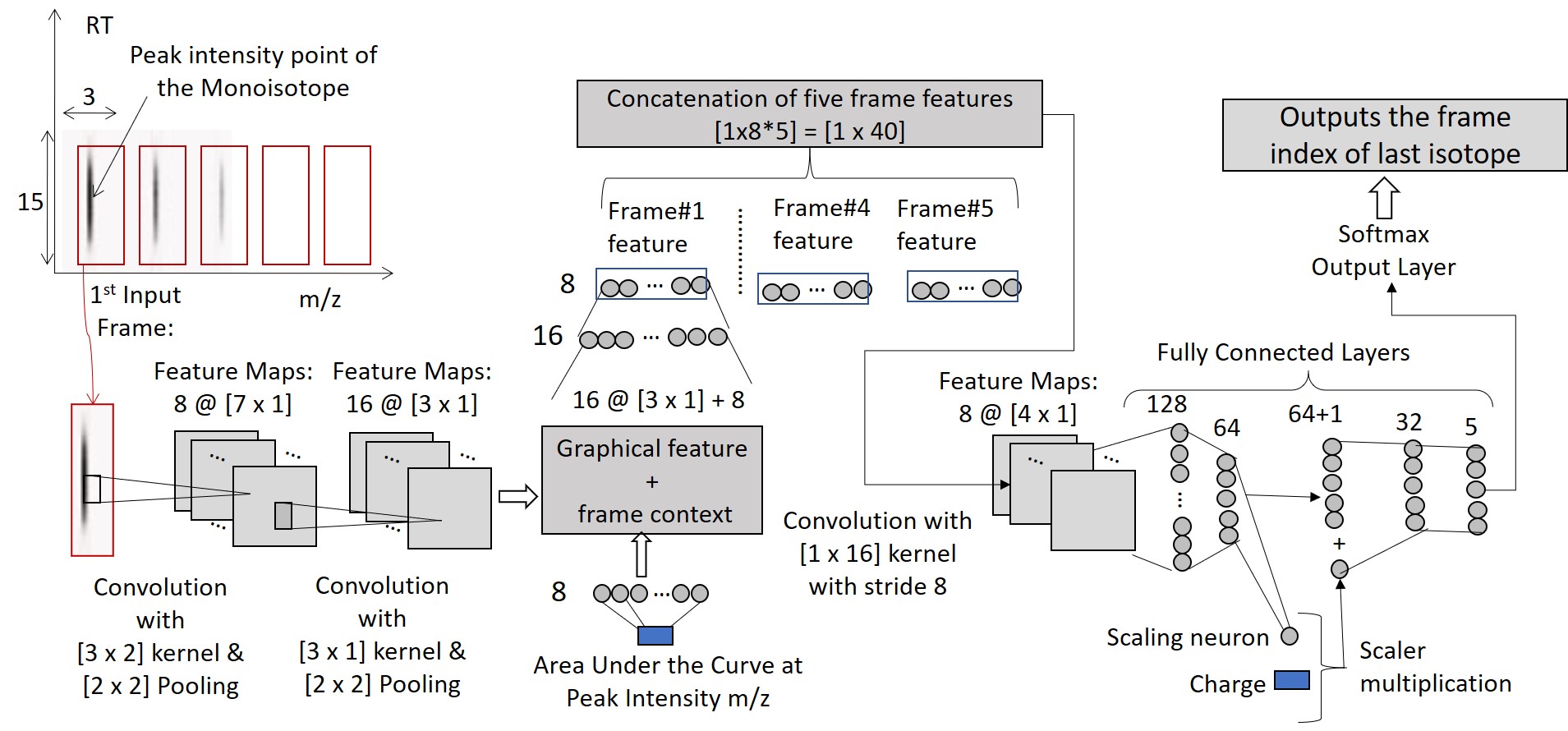}
\caption{Network of IsoGrouping module. It starts with two convolution layers to fetch the graphical features from the input frame. Then we concatenate the total intensity (AUC) of the isotopic signal  with it through an embedding layer of neurons (frame context). Then this is passed through two fully connected layers having size 16 and 8. This gives us `frame feature' of the input frame. We perform the same for five consecutive frames and then concatenate the `frame feature' of those altogether. Then one layer of convolution is applied to detect combined feature from all the frames. The resultant features are passed through two fully connected layers (size 128 and 64) to decide whether this is a noise or potential feature. This probability is also used to activate a scaling neuron, that feeds the charge into the network through proper scaling. Scaled charge is  concatenated with the latest layer output (size 64) and passed through two fully connected layers. Finally the Softmax output layer at the end classifies the sequence. We include pooling layers after first and second convolution layers. We apply ReLu activation function for the neurons. The dropout layers are included after each fully connected layers with dropout probability of $0.5$. The other network parameters are mentioned in the figure.}\label{isogrp}
\end{figure}

IsoDetecting module provides us a list of isotopes. Equidistant isotopes having same charge are grouped into a cluster or sequence. Then those sequences of isotopes are passed to the IsoGrouping module. Unlike DeepIso, we do not pass the isotopes directly to the second module, because here the input resolution is different in two modules. Therefore we apply a resolution degradation technique that filters out the region of isotope (as suggested by IsoDetecting module) from the background, presents it in lower resolution, and then pass it to the IsoGrouping module. The detailed procedure is provided in the Supplementary Method B.

Our proposed network for IsoGrouping module is illustrated in Figure~\ref{isogrp}. It may break each sequence into multiple features, or report one feature consisting of the whole sequence of isotopes, or even report that sequence as merely noise. It works on a sequence of isotopes in multiple rounds. Each round process five consecutive isotopes at a time. The output is $i=0$ to $4$, where, $i=0$ means that no feature starts at the first frame, so skip it. Output $i=1$ to 4 means, there is a feature starting in the first frame, and it ends at $(i+1)^{th}$ frame. If output $i=4$, it means that the feature might have more than 5 isotopes. Those can be found by overlapping rounds. A step by step explanation of the scanning procedure with figure is provided in Supplementary Method C.  

%\begin{equation}
%f_t = (1-a_t).f_{t-1} + a_t.f'_t   % \end{equation}

%\begin{align}
%f'_t &=H(W_{hh}.f_{t-1}+ W_{oh}.X_{ot} + b_h) \\
%a_t &=\sigma(W_a.f'_t + b_a)     
%\end{align}
%n Equation 3, $H$ is the activation function, $W_{hh}$ is the weight matrix connecting the previous hidden state $f_{t-1}$ to the current state, $X_{ot}$ is the output of the layer $o$, $W_{oh}$ is the weight matrix connecting the $X_{ot}$ to the RNN layer, $b_h$ is the bias at the RNN layer. In Equation 4, $\sigma$ is the sigmoid activation function, $W_a$ is the weight matrix that learns the attention mechanism and $b_a$ is the corresponding bias. % Equation 2 is doing element-wise multiplication, whereas, Equation 3 and 4 are doing matrix-multiplications.  

\subsubsection*{Training Procedure of IsoGrouping module}
Usually monoisotope's intensity is the highest among the other isotopes in a feature and dominates the total intensity of the feature. This property should be learnt by IsoGrouping module. %For instance, isotope 1 listed in the feature table shown in Figure~\ref{alg}. Our network should learn to detect mono isotopes.
We prepare the positive samples by generating a sequence of 5 frames for each peptide feature, where the sequence starts at the first isotope of the respective feature. Each frame has dimension $[15x3]$, covering 15 scans along RT axis and 3 units along the $m/z$ axis. We filter out the isotopic signal from background by taking the intensity within range 2 ppm before and after the peak intensity $m/z$ value, and 7 scans before and after the peak intensity RT value. The signal is left aligned with the frame. Each sequence is labeled by the frame index holding the last isotope of the feature (indexing starts from 0). Minimum number of isotopes in a feature is 2 , i.e., label is 1. If the feature has equal of more than 5 isotopes, label is 4. 
%In this way we generate about 220,00 positive samples.
We generate negative samples by cutting some sequences from the noisy or blank area. We also generate sequences that contain peptide feature, but the feature does not start at the first frame of the sequence. Those samples are labeled as `0' as well. We do this to handle the cases where noisy traces are classified as isotopes by IsoDetecting module by mistake and thus clustered with the actual features in the intermediate step. We see the training and validation sensitivity in Table~\ref{train_val_2}. 
For the training, we set minibatch size 128 and apply `Adagrad' stochastic optimization ~\cite{duchi2011adaptive} with initial learning rate of $0.07$. We use softmax cross entropy as error function at the output layer. 
 
\begin{table}[ht]
\centering
%\resizebox{\textwidth}{!}{
\begin{tabular}{|c|c|c|}
\hline
Class & Sensitivity on Training Set (\%) & Sensitivity on Validation Set (\%)\\
\hline
0 (noise) & 89.42 & 90.78 \\
\hline
1 (2 isotopes) & 57.93 & 57.50\\
\hline
2 (3 isotopes) & 51.98 & 43.30\\
\hline
3 (4 isotopes) & 61.90 & 59.86\\
\hline
4 (5 isotopes or more) & 61.77 & 64.14\\
\hline
\end{tabular}
%}
\caption{\label{train_val_2}Class sensitivity of IsoGrouping module on training set and validation set. The output is $i=0$ to $4$, where, $i=0$ means that no feature starts in the first frame, so skip it. Output $i=1$ to 4 means, there is a feature starting in the first frame, and it ends at $(i+1)^{th}$ frame. When output $i=4$, it means there might be more isotopes left. So we run another round of processing over the rest of the isotopes of the same cluster or sequence.  So although our network process 5 frames at a time, but if the feature has more than 5 isotopes, those can be found by overlapping rounds. }
\end{table}

We observe that the maximum sensitivity of the classes is about 60\%. To have a better perception we present the confusion matrix in Table~\ref{conf_mat_2}. We see the model hardly misses the monoisotopes (low percentage of features misclassified as class A), but confuses about the last isotope of a peptide feature. Please note that reporting the monoisotope along with first few isotopes (having higher intensity peaks) of a feature is more important in the workflow. Because they dominate the feature intensity and used in the next steps of protein quantification and identification. 
\begin{table}[ht]
\centering
%\resizebox{\textwidth}{!}{
\begin{tabular}{|c|c|c|c|c|c|}
\hline
Class & 0 & 1 & 2 & 3 & 4 \\
\hline
0 & 89.43\% & 6.88\% & 1.72\% & 0.94\%& 1.03\%\\
\hline
1 & 17.29\% & 57.93\% & 18.18\% & 5.58\% & 1.03\% \\
\hline
2 & 5.38\% & 18.94\% & 51.98\% & 21.58\% & 2.13\%\\
\hline
3 & 3.25\% & 5.44\% & 14.29\% & 61.90\% & 15.12\% \\
\hline
4 & 5.71\% & 2.55\% & 3.18\% & 26.79\% & 61.77\% \\
\hline
\end{tabular}
%}
\caption{\label{conf_mat_2}Confusion matrix produced by IsoGrouping module on validation dataset. The diagonal values, e.g. [2, 2] represent the sensitivity for class 2. We say a feature is misclassified as class 0 when the monoisotope (first isotope) or all of the isotopes are missed, i.e., the feature is thought to be noise by mistake. The value of [2, 0] indicates what percentage of features with three isotopes are either misclassified as noise, or monoisotope is missed. [2, 1] indicates the percentage of features which actually have three isotopes but the third one is missed, and only first two are combined together. Similarly [2,3] shows for what percentage of three isotope features, IsoGrouping module finds ONE additional isotope at the end.}
\end{table}

Finally, we would like to mention some common strategies followed for implementing and training both of the modules. We implemented our deep learning model using the Google developed Tensorflow library. %Number of filters and filter size are decided based on experiments.
We check the accuracy on validation set after training on every 1200 samples. We perform data shuffling after each epoch which helps to achieve convergence faster. We continue training until no progress is seen on validation set for about 15 epochs. For developing the PointIso we use Intel(R) Xeon(R) Gold 6134 CPU, NVIDIA Tesla GPU, and Ubuntu 16.04.5 LTS operating system. %Including dropout layer in our model increases the validation sensitivity by about 1.5\%. Although the Rectifier Linear Unit (ReLu) activation function is preferred over tanh in many literature, our model does not learn well with ReLu according to our experiments. 

\section*{Data Availability}
The benchmark dataset is available to download from ProteomeXchange using accession number PXD001091. The full experimental result on all the replicates of the samples are available in supplementary materials. Source code can be found at this address: https://github.com/anne04/PointIso

\bibliography{sample}

%\noindent LaTeX formats citations and references automatically using the bibliography records in your .bib file, which you can edit via the project menu. Use the cite command for an inline citation, e.g.  \cite{Hao:gidmaps:2014}.

%For data citations of datasets uploaded to e.g. \emph{figshare}, please use the \verb|howpublished| option in the bib entry to specify the platform and the link, as in the \verb|Hao:gidmaps:2014| example in the sample bibliography file.

%\section*{Acknowledgements (not compulsory)}
%Acknowledgements should be brief, and should not include thanks to anonymous referees and %editors, or effusive comments. Grant or contribution numbers may be acknowledged.

\section*{Acknowledgements}
This work is partially supported by NSERC OGP0046506, NSF China grant 61832019, Canada Research Chair Program, The National Key R\&D Program of China 2018YFB1003202, and Bioinformatics Solutions Inc. We thank Rui Qiao, Department of Statistics and Actuarial Science, University of Waterloo, Waterloo, ON, Canada, for sharing his thoughtful ideas about PointNet and class imbalance problems.

\section*{Author contributions}
F.T.Z designed and developed the model, and performed the experiments as well. M.Z.R helped with producing database search result by MASCOT and setting parameters of other existing tools. M.Z.R. and L.X. helped with studying the  characteristics of peptide features and LC-MS map. N.H.T contributed by suggesting various deep learning ideas. B.S. and M.L. proposed and supervised the project. All authors reviewed the manuscript. 

\section*{Competing interests}
The authors declare no competing interests.
\newpage

%\section*{Additional information}
%To include, in this order: \textbf{Accession codes} (where applicable); \textbf{Competing interests} (mandatory statement). 

%\includepdf[]{deepIsoV2_final.pdf} 

\captionsetup[table]{name=Supplementary Table S}
\captionsetup[figure]{name=Supplementary Figure S}

%\counterwithin{figure}{section}
%\counterwithin{table}{section}

\section*{PointIso: Point Cloud Based Deep Learning Model for Detecting Arbitrary-Precision Peptide Features in LC-MS Map through Attention
Based Segmentation}

Fatema Tuz Zohora, M Ziaur Rahman, Ngoc Hieu Tran, Lei Xin, Baozhen Shan, Ming Li

%\section*{Physiochemical Property of Features Detected by Different Tools}

\section*{Supplementary Note A} 
%\subsection*{Class sensitivity of IsoDetecting module for different weighting mechanisms}
\begin{table}[ht]
\centering
%\resizebox{\textwidth}{!}{
\begin{tabular}{|c|c|c|c|c|c|c|}
\hline
Experiment Category & z=0 & z=1 & z=2 & z=3 & z=4 \\
\hline
Without any class weight (after 10 epochs) & 96\% & 40\% & 53\% & 33\% & 15\%  \\
\hline
\thead{Class weight based on distribution \\over whole dataset (after 10 epochs)} & 91\% & 48\% & 68\% & 51\% & 27\% \\
\hline
\thead{Class weight based on distribution \\over sample (after 10 epochs)} & 86\% & 51\% & 71\% & 55\% & 30\% \\
\hline
Same as above, after convergence & 64\% & 59\% & 79\% & 67\% & 14\% \\
\hline
Same as above, with attention mechanism & 50\% & 79\% & 95\% & 91\% & 85\% \\
\hline
\thead{Same as above, with attention mechanism \\and higher resolution} & 40\% & 99\% & 98\% & 98\% & 98\% \\
\hline
\end{tabular}
%}
\caption{\label{weight} Class sensitivity for different weighting mechanisms. We compare the candidate weighting mechanisms based on the sensitivity for the best case scenario, i.e., when feature is aligned with the left boundary of the scanning window (e.g., feature A in Figure 4(a)),  with high abundant features, i.e., features having charge, z = 1, 2, 3, 4. Please note that, although the sensitivity of negative class (z=0) is comparatively lower for our chosen criteria, however, it does not imply that it reports many false positives. Although the datapoints which are very close or adjacent to the real signal, are sometimes predicted as positive points, but in general the negative class has higher class sensitivity than all others as presented in Table 7 of the main manuscript.}
\end{table}
\subsection*{Class Weight Assignment Procedure}
We choose the class weight based on distribution per sample. Let us have a sample window which contains 100 datapoints. They come from three different classes ($z$): 0, 2, and 3. We have 10 points from $z=3$, 30 points from $z=2$, and remaining $60$ points from $z=0$. Then points from class 0, 2, and 3 get weights of 0.4, 0.7, and 0.9 respectively. We perform this for every sample. So every training sample has a weight list associated with it, which we pass to the network during cross entropy loss calculation.  

\section*{Supplementary Note B} 
\subsection*{Attention Mechanism}
We would like to discuss our experiments to correctly segment partially covered peptide features in a target window. We started with considering a bigger scanning window which will produce output for the center region. This caused three problems. First, we can apply back-propagation based on full bigger window during training time (number of input nodes equals to the number of output nodes), but use the center region only during testing time or prediction (output nodes corresponding to center region are used only). In this case model gets confused about the boundary region during training and cannot learn well. Second, during training time we can apply back-propagation based on center region only, therefore, number of input nodes is not equal to the number of output nodes anymore. This actually makes the architecture very complex. Because number of datapoints in the center region, and surrounding r1, r2, r3, r4 regions is never fixed. Although we consider a fixed number and do padding for simplification, but when number of input is not equal to the number of outputs, internal design gets quite complicated. Also the fact that, datapoints in r1, r2, r3, r4 need not to worry about each other, only have to focus on center region, may not be well learned by the model. We would not have any control over that. Finally, we also face technical issues since GPU memory exceeded (more than 16 GB) with bigger region. Because of these reasons we were unable to proceed with  the design. %, although it seemed simple at the beginning.

\begin{table}[ht]
\centering
%\resizebox{\textwidth}{!}{
\begin{tabular}{|c|c|c|c|c|c|c|}
\hline
Experiment Category & z=0 & z=1 & z=2 & z=3 & z=4 \\
\hline
Sliding window with 50\% overlapping & 86\% & 50\% & 72\% & 57\% & 36\%  \\
\hline
Skiplink inserted in above model & 85\% & 56\% & 76\% & 66\% & 41\% \\
\hline
Bi-directional 2D RNN & 85\% & 51\% & 77\% & 67\% & 49\% \\
\hline
\textbf{Attention mechanism} & 84\% & 62\% & 85\% & 71\% & 50\% \\
\hline
\textbf{Attention mechanism with higher resolution} & 90\% & 64\% & 85\% & 81\% & 61\% \\
\hline
\end{tabular}
%}
\caption{\label{s_atn_table} Different techniques of absorbing surrounding information and corresponding class sensitivity of IsoDetecting module. We define the class sensitivity of a scanning window as the number of datapoints from class $z$ (0 to 9) detected correctly out of total number of datapoints in a scanning window. To evaluate candidate solutions we use
the class sensitivity of high abundant features (charge $z=$ 1,2,3, and 4) in a 
\emph{average case} scenario. Average case means the scanning window might contain any number of features, they may appear at any location of the window, they might be partially or fully seen, and might be overlapping as well. We see that the DANet inspired attention based mechanism works better than other techniques.}
\end{table}

So we applied next simpler technique, sliding windows with 50\% overlapping. The resultant class sensitivities are presented in the first row of Table~S\ref{s_atn_table}. After passing the IsoDetecting output through IsoGrouping module, it finally produce only 65\% feature detection as presented in the first row of Table 5. Therefore we had to use more sophisticated approach to address this problem. 

In Figure 4 of main manuscript, we see that the regions $r_1$, $r_2$, $r_3$, and $r_4$ are actually playing the key role in detecting the traces inside target window. We empirically tested following three techniques for diffusing the surrounding information into current window, and the results are summarized in Table~S\ref{s_atn_table}. 
\begin{itemize}
    \item First, we just calculated the global features of surrounding regions and diffuse them together by addition and concatenation with the global features of target window. Then we repeated the 50\% overlapping technique, which did not bring any significant change. Using some skip links along with that brings little improvement as shown in the second row of the table. 
    \item Then we used a bi-directional two dimensional RNN network to flow information from all direction into the target window. The corresponding class sensitivities are provided in the third row. %It finally results in 72\% feature detection as shown in the second row of Table~\ref{progress}.
    \item   Finally, we applied the attention mechanism proposed by DANet which works better than first two techniques, as reported in the fourth and fifth row. So we choose PointNet segmentation network combined with DANet to develop our IsoDetecting module. 
\end{itemize}

\section*{Supplementary Note C} 

\subsection*{Controlling False Detection Rate}
The secondary signals very close to the primary signals are usually merged with the primary signal through some pre-processing which involves different parameter settings based on dataset and context. However, we tend to avoid all kind of heuristic parameters and therefore work on raw LC-MS map. That is why our algorithm has to perform noise removal as well as feature detection. Although random noise removal is learned easily by PointIso but separation of feature like noisy traces and those secondary peaks removal is difficult for PointIso. This tasks has to be performed by IsoDetecting module since it has the access to whole context. So it has to see through all the signals and decide which ones are to be reported and which are to be ignored. Our initial trained model reports those feature like signals as peptide features. Therefore we fine tune the model to resolve this problem.

In order to do collect those wrong reports, we first let our model scan through a LC-MS map and report peptide feature list. Then we match that list with the peptide feature list produced by all other tools (i.e., OpenMS, MaxQuant, Dinosaurs, and PEAKS) with an error tolerance of $0.01 m/z$ and $0.2$ min RT. The features from PointIso list which did not match are selected for fine tuning (about 70,000 features). We cut those features' corresponding scanning window along with surrounding region, as done for other training samples. Then we start with already trained model,  run few epochs with 0.0001 learning rate and keep the best model state. %saver.save(sess, modelpath+log_no+'_best_model_1.ckpt')
We use about 30,000 samples for fine tuning. We can balance between the true detection rate and false detection rate using the amount of these samples. 

This same set of features can be also used for fine tuning IsoGrouping module. But its not very effective since it sees only filtered out signals and does not have access to the bigger context like IsoDetecting module. We save two types of model state while training the IsoGrouping module. One is best loss model, and another is best sensitivity model based on validation dataset. Best loss model gives us lower number of peptide features then the best sensitivity model, however, detection percentage is also dropped a little bit. For example, we get 97.53\% detection with 80,000 features, instead of 98.06\% detection with 100,000 features (which is reported in the main result). Without any fine tuning we get 99.50\% detection, with about 200,000 features. 

In our manuscript we have used fine tuned model of IsoDetecting module, and best sensitivity model of IsoGrouping module. But the other models are  also uploaded in the GitHub repository. Users can choose according to their need. They can also fine tune further if necessary. 

\section*{Supplementary Method A}

\subsection*{Attention Calculation}
\begin{figure}[h!]
\centering
\includegraphics[scale = .5]{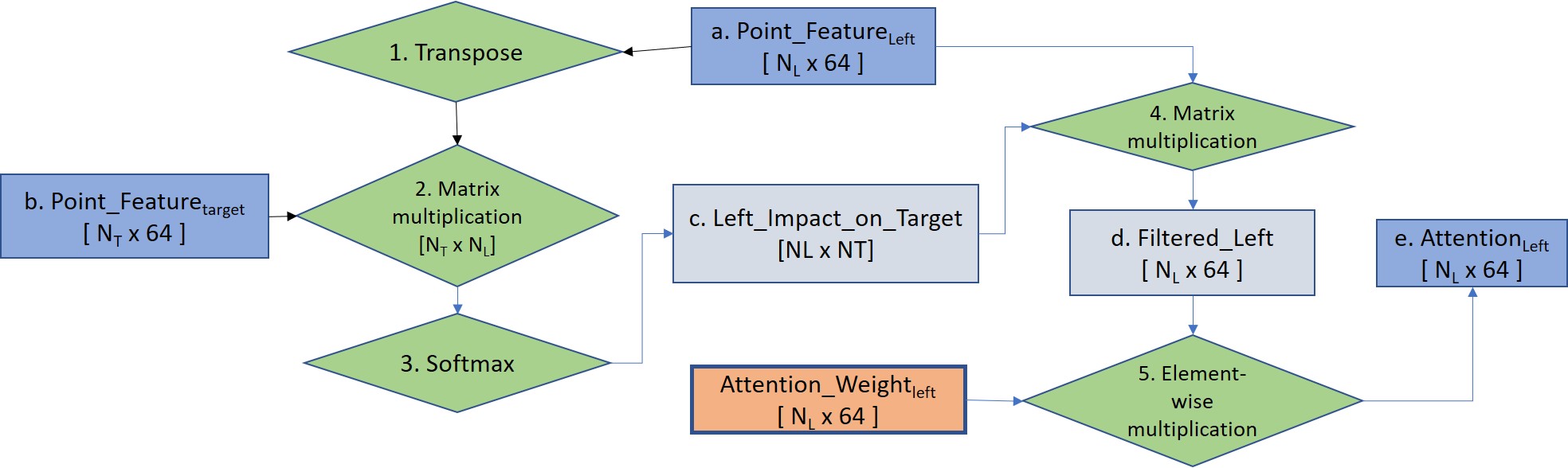}
\caption{Flowchart of attention calculation in IsoDetecting module. This particular flowchart is intended to find out the attention or impact of left region over the datapoints of target window. Exactly similar approach is followed for other surrounding regions as well and finally all are diffused with the $point\_features\_target$ by addition.}\label{s_danet}
\end{figure}

Please refer to the flowchart shown in Figure~\ref{s_danet}. 
We have the point features of left region, $point\_feature\_left$ according to the PointNet architecture shown in the main manuscript. Then we multiply $point\_feature\_target$ with the transpose of $point\_feature\_left$ according to the rule of matrix multiplication. Then we take softmax of the product as Equation 1, which gives us a $[N_T \times N_L]$ matrix, $left\_impact\_on\_target$, where, $N_T$ and $N_L$ are the total number of datapoints of the target window and left region respectively. So each row presents a datapoint from target window and columns present the datapoints from left region.

\begin{equation}
left\_impact\_on\_target =
\frac{exp( (point\_feature\_target_i) . (point\_feature\_left_{j}))}{\sum_{\text{$i = 1$ to $N_L$}} \text{ } exp((point\_feature\_target_i) . (point\_feature\_left_j))}
\end{equation}
Therefore, 
$left\_impact\_on\_target_{(j,i)}$ presents the attention of $i^{th}$ point of left region over the $j^{th}$ point of target window. The higher the value, the higher the correlation (similar feature) between those two points. So the $j^{th}$ row tells us which datapoints from left region have higher attention or highly correlated with the $j^{th}$ datapoint of target window. Next, we want to fetch those *significant point features from left region. So we apply another round of matrix multiplication ($4^{th}$ operation) between $left\_impact\_on\_target$ and $point\_feature\_left$. We denote the resultant product as $filtered\_left$ since it essentially gives us $point\_features\_left$ but scaled/filtered according to the aforementioned correlation or attention. Then again we have to know how much of those filtered features should be incorporated with the $point\_features\_target$ while segmenting the datapoints of target window. So we use a weight matrices, namely $Attention\_Weight\_left$,  and multiply it with $filtered\_left$, producing $attention\_left$, which is finally passed forward to be diffused (by addition) with the $point\_features\_target$. This $Attention\_Weight\_left$ is learned through training.

\section*{Supplementary Note D}

\subsection*{Basic Properties of Peptide Feature}

It is supposed to learn following basic properties of peptide feature, besides many other hidden characteristics from the training data.
\begin{enumerate}
\item In the LC-MS map, the isotopes in a peptide feature are equidistant along $m/z$ axis. For charge $z=$ 1 to 9, the isotopes are respectively 1.00 $m/z$, 0.5 $m/z$, 0.33 $m/z$, 0.25 $m/z$, 0.17 $m/z$, 0.14 $m/z$, 0.13 $m/z$, and 0.19 $m/z$ distance apart from each other~\cite{steen2004abc}.
\item The intensities of the isotopes form bell shape within their retention time (RT) range as shown in the zoomed in view of Figure 1 of the main manuscript. 
\item Peptide features often overlap with each other. 
\end{enumerate}

\section*{Supplementary Method B}
\subsection*{Resolution Degradation in IsoGrouping Module}
IsoDetecting module generates sequences of potential isotopes which are sent to the IsoGrouping module for final  detection of peptide features. Now, the $m/z$ value of isotopic signals are real numbers having up to 4 decimal places. We degrade each signal into real numbers having up to 2 decimal places. For example, let us have two sequences of isotopes $A$, and $B$, where the first isotope of the sequences are denoted as $A_1$ and $B_1$ respectively. The $m/z$ values of these are respectively $A_{1\_mz}=500.2351$ and $B_{1\_mz}=500.2443$. During resolution degradation we filter out the signal instensity from the background with +-2 ppm $m/z$ range. The range is calculated as:$\frac{A_{1\_mz}\times 2.0}{10^6}$. So that we have $A_{1\_mz}=500.24$ and intensity is set as the maximum of the intensities of datapoints within range $500.2341$ to $500.2361$. Similarly, $B_{1\_mz}=500.24$, \emph{but}  intensity is set as the maximum of the intensities of datapoints within range $500.2433$ to $500.2453$. So we see that, although both have the same $m/z$ values in lower resolution, they definitely belong to different sequences and have different intensities (thus might have different pattern as well). 

\section*{Supplementary Method C}
\subsection*{Scanning Procedure by IsoGrouping Module}
This method is intentionally kept similar to DeepIso. Although we have upgraded the internal architecture significantly, for the sake of user convenience we do not change the scanning procedure. Therefore, we request our readers to refer to the corresponding material from DeepIso.

\begin{figure}[h!]
\centering
\includegraphics[scale = .55]{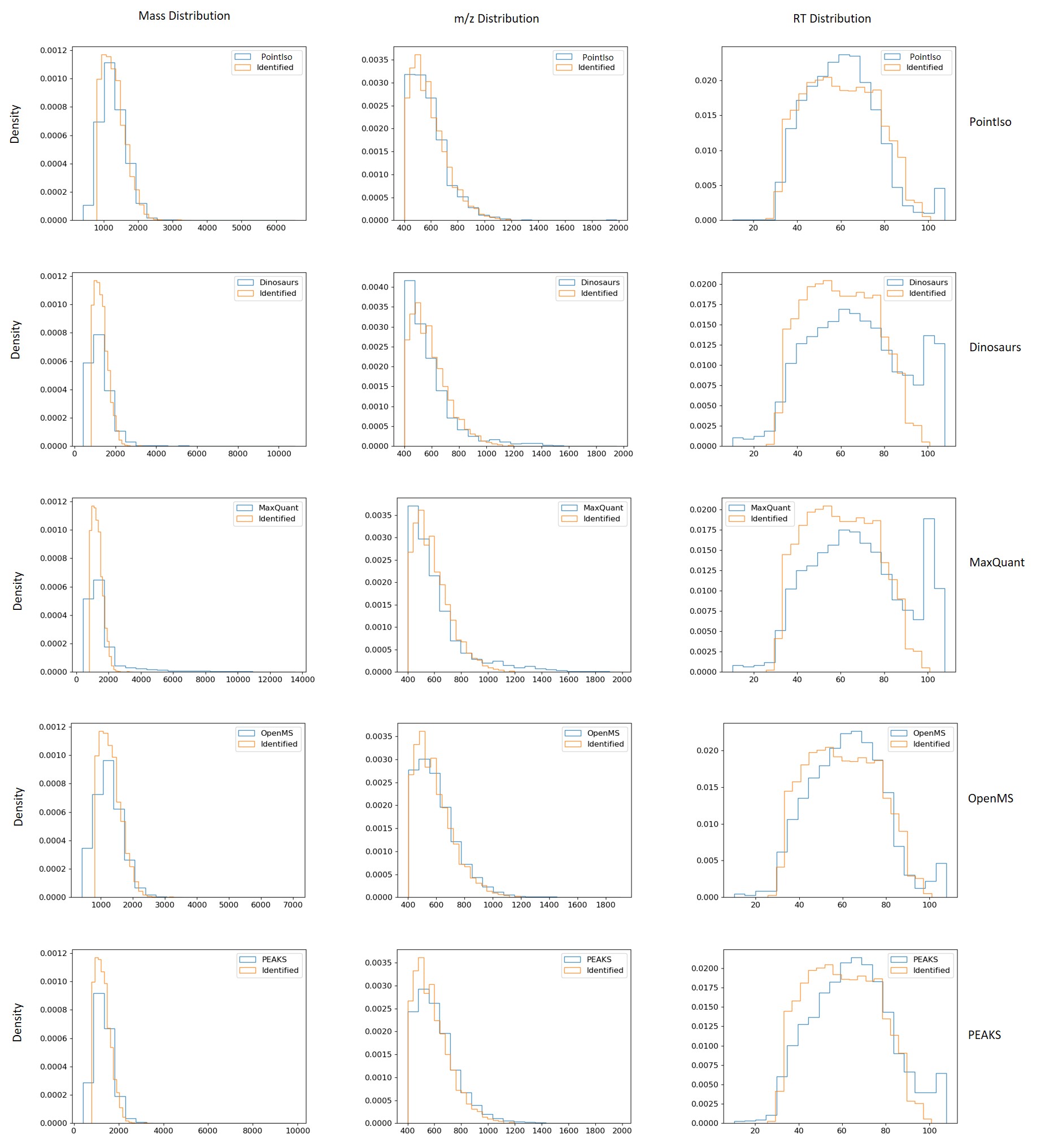}
\caption{Comparison of mass, m/z and RT distribution of detected features (blue) and identified features (orange) for different tools.}\label{s_dist}
\end{figure}

%\section*{Supplementary Table S3}
\includepdf[pages=-]{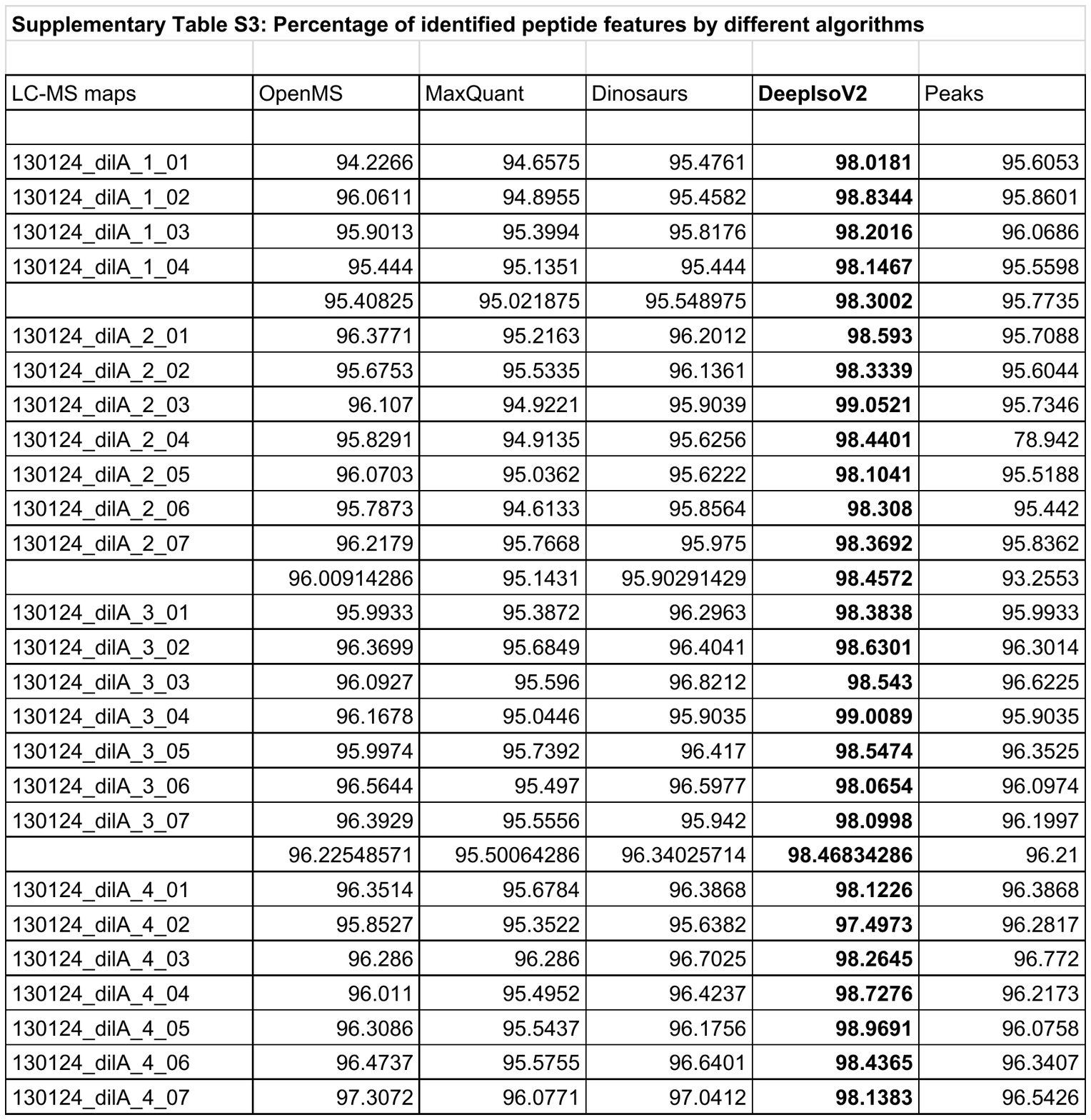}

\end{document}